\documentclass[A4]{IEEEtran}
\IEEEoverridecommandlockouts
\usepackage{cite}
\usepackage{amsmath,amssymb,amsfonts}
\usepackage{amssymb}
\usepackage{breqn}
\usepackage{cleveref}
\usepackage[lined,ruled,linesnumbered,commentsnumbered]{algorithm2e}
\usepackage{booktabs}
\usepackage{cases}
\usepackage{graphicx}
\usepackage{multirow}
\usepackage{mathrsfs}
\usepackage{textcomp}
\usepackage{xcolor}
\usepackage{subfigure}

\newtheorem{myDef}{Definition}

\def\BibTeX{{\rm B\kern-.05em{\sc i\kern-.025em b}\kern-.08em
    T\kern-.1667em\lower.7ex\hbox{E}\kern-.125emX}}

\begin{document} 

\title{Manifold Interpolation for Large-Scale Multi-Objective Optimization via Generative Adversarial Networks}

 \author{
	Zhenzhong~WANG,~\IEEEmembership{}
	Haokai HONG,~\IEEEmembership{}
	Kai YE,~\IEEEmembership{}
	Min JIANG$^\star$,~\IEEEmembership{Senior~Member,~IEEE,}
	and Kay Chen TAN,~\IEEEmembership{Fellow,~IEEE}
	\IEEEcompsocitemizethanks{\IEEEcompsocthanksitem Z. WANG, H. HONG, K. YE, and M. JIANG are with the School of Informatics, Xiamen University, China, Fujian, 361005. \protect M. JIANG is the corresponding author (Email: minjiang@xmu.edu.cn).
		\IEEEcompsocthanksitem {Kay Chen TAN is with the Department of Computer Science, City University of Hong Kong, and the City University of Hong Kong Shenzhen Research Institute.}
	}	
}

\bibliographystyle{IEEEtran}
\maketitle

\begin{abstract}
Large-scale multiobjective optimization problems (LSMOPs) are characterized as involving hundreds or even thousands of decision variables and multiple conflicting objectives. An excellent algorithm for solving LSMOPs should find Pareto-optimal solutions with diversity and escape from local optima in the large-scale search space. Previous research has shown that these optimal solutions are uniformly distributed on the manifold structure in the low-dimensional space. However, traditional evolutionary algorithms for solving LSMOPs have some deficiencies in dealing with this structural manifold, resulting in poor diversity, local optima and inefficient searches. In this work, a generative adversarial network (GAN)-based manifold interpolation framework is proposed to learn the manifold and generate high-quality solutions on this manifold, thereby improving the performance of evolutionary algorithms. We compare the proposed algorithm with several state-of-the-art algorithms on large-scale multiobjective benchmark functions. Experimental results have demonstrated the significant improvements achieved by this framework in solving LSMOPs.





\end{abstract}

\begin{IEEEkeywords}
evolutionary algorithm, multiobjective optimization, large-scale optimization, generative adversarial networks, manifold learning
\end{IEEEkeywords}

\section{INTRODUCTION}
   
Many real-world optimization problems involve hundreds or even thousands of decision variables and multiple conflicting objectives, and such problems are called large-scale multiobjective optimization problems (LSMOPs) \cite{6191315 ,Stanko2015Large,9185798}. For example, in the design of telecommunication networks\cite{208913}, several tens of thousands of variables, such as the locations of many network nodes, the transmission capacity between nodes, and the power allocations of nodes, are involved. These numerous decision variables determine the energy consumption, applicability, and stability of the network. It is a challenge to find optimal solutions for this type of network because the volume of the search space is exponentially related to the number of decision variables, and thus the curse of dimensionality \cite{wang2015memetic} is exhibited. Therefore, an excellent large-scale multiobjective optimization algorithm (LSMOA) should overcome this issue to search in the search space efficiently and effectively, and this ability is crucial to solving LSMOPs.

To solve LSMOPs, a variety of LSMOAs have been proposed in recent years. These LSMOAs can be roughly grouped into the following three categories. The first category of LSMOAs is based on decision variable analysis. A representative algorithm of this category is MOEA/DVA \cite{7155533}, where the original LSMOP is decomposed into a number of simpler subproblems. Then, the decision variables in each subproblem are optimized as an independent subcomponent. The second category of LSMOAs is based on decision variable grouping. This category of LSMOAs divides the decision variables into several groups and then optimizes each group of decision variables successively. For example, C-CGDE3 \cite{6557903} maintains several independent subpopulations. Each subpopulation is a subset of the equal-length decision variables obtained by variable grouping. 
The above decision variable analysis or grouping methods easily leads to excessive computational complexity due to the large-scale decision variables \cite{hongye111}.
The third category is based on problem transformation. A generic framework named the large-scale multiobjective optimization framework (LSMOF) \cite{8628263} is representative of this category. In LSMOF, the original LSMOP is reformulated as a low-dimensional single-objective optimization problem with some direction vectors and weight variables, aimed at guiding the population towards the optimal solutions.

\begin{figure}[htbp] 
  \centering   
  \includegraphics[width=8.88cm]{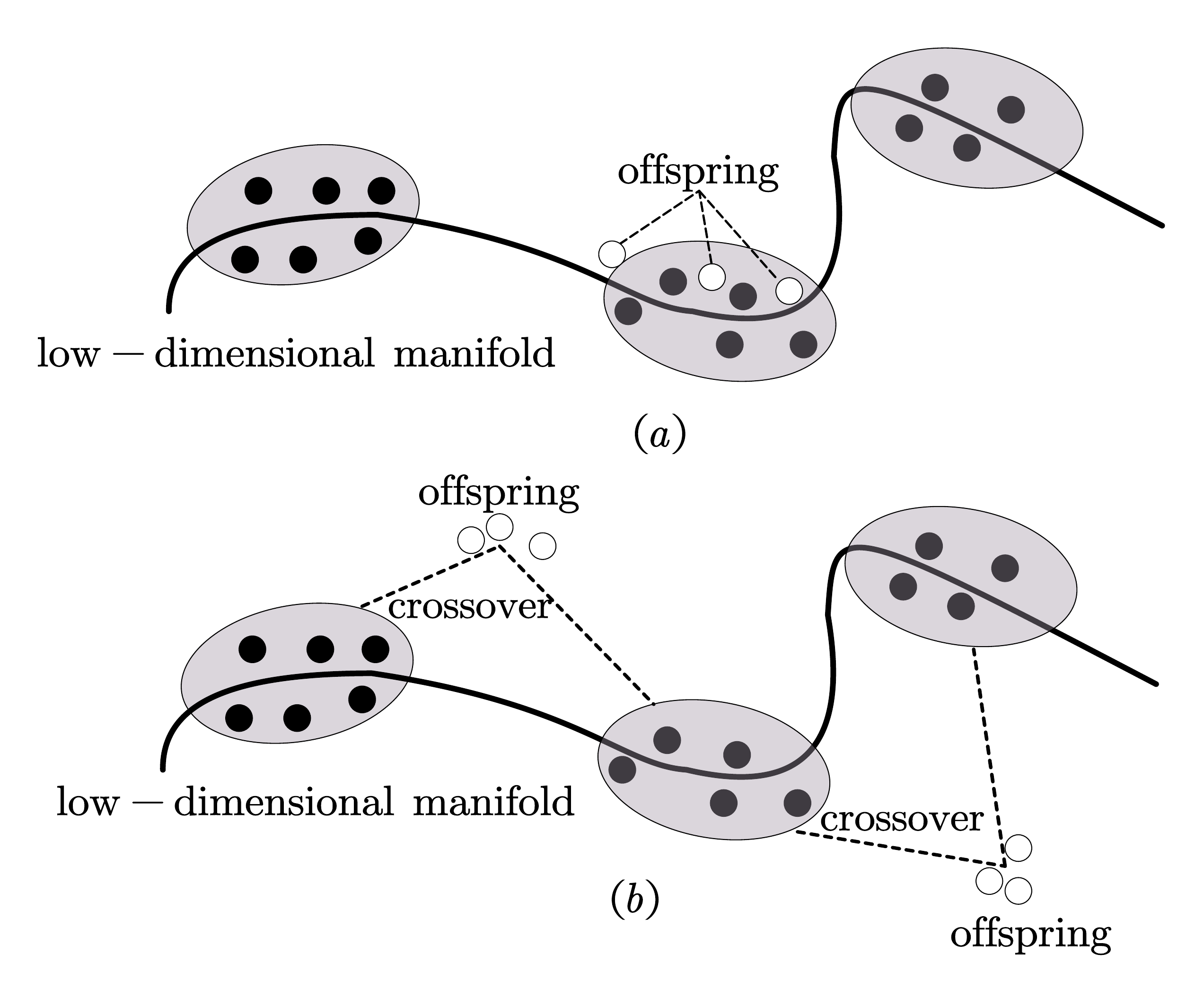}
  \caption{Solutions are piecewise on this low-dimensional manifold, which results in poor diversity, local optima and inefficient search.}
  \label{fig:idea}
\end{figure}

Recently, research has shown that the Pareto-optimal solutions are uniformly distributed on the manifold structure in the low-dimensional space \cite{Hillermeier2001Nonlinear,Mardle1999Nonlinear}, and utilizing such manifold can be an efficient way to improve the evolutionary performance \cite{4358761,7155533,min2020EvolutionaryManifold,9185009}. However, traditional algorithms for solving LSMOPs have some deficiencies in dealing with this structural manifold, which may blight the manifold structure and result in poor diversity, local optima and inefficient search. For example, as shown in Fig. \ref{fig:idea} (a), the solutions are piecewise on the manifold due to the large-scale search space. Solutions from the same segment that have similar genes are mutated or mated, and the offspring solutions are likely to lie around the segment. In this case, the diversity of the population cannot be improved significantly, and it is disadvantageous to escape from local optima. Alternately, as shown in Fig. \ref{fig:idea} (b), solutions from different segments that have dissimilar genes are mated, the diversity of the population can be improved, and the offspring solutions are likely to escape from local optima, but these offspring solutions could be made obsolete by nondominated sorting, which leads to inefficient search.

\begin{figure}[htbp] 
  \centering   
  \includegraphics[width=8.88cm]{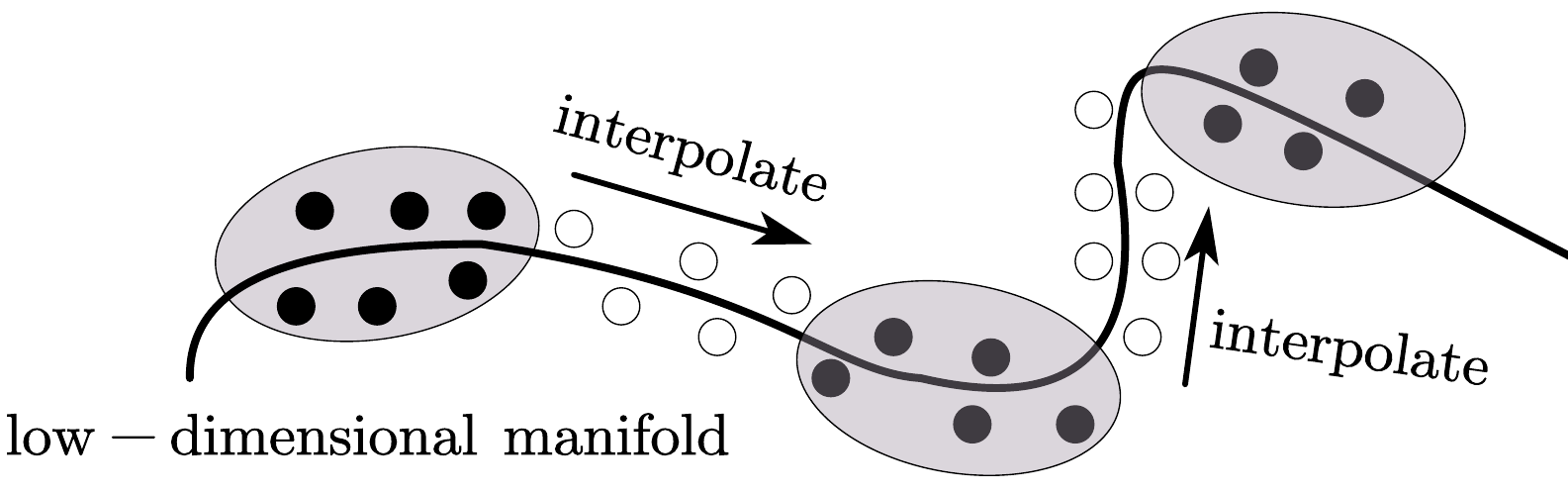}
  \caption{Interpolate solutions on the manifold}
  \label{fig:idea2}
\end{figure}

We believe that the key factor for solving LSMOPs is to find an effective way to learn the manifold and search for solutions on this manifold. An intuitive idea is to learn the manifold and interpolate solutions along the manifold, as shown in Fig. \ref{fig:idea2}. These interpolated solutions fill gaps of the manifold and are uniformly distributed on the manifold, thereby leading the population to draw on the optimal area efficiently. Unfortunately, it is difficult to learn the characteristics of the entire manifold and fill its gaps because we can only learn characteristics from existing solutions. For solutions lying in the gaps that have not yet been found, their characteristics cannot be learned directly.

Development from the neural network community has already shown that a group of generative models, GANs \cite{NIPS2014_5352}, provides a powerful tool for learning the high-dimensional characteristics of samples and further generating meaningful samples. From the perspective of manifold data, GANs can learn the latent characteristics lying on the manifold and interpolate previously nonexistent but meaningful samples located on this manifold \cite{NIPS2018_7964,8627945,NIPS2014_5423}. By this interpolation method, a series of samples with continuous changes are synthesized on the manifold, and thus the gaps of the manifold can be fulfilled, and we can utilize the knowledge of the manifold to guide the search direction. Therefore, the GAN can be utilized as a powerful weapon in our work to interpolate solutions on the manifold.

In this paper, we argue that the integration of GANs into multiobjective optimization algorithms can offer significant benefits for designing better LSMOAs, and a framework called GAN-based large-scale multiobjective evolutionary framework (GAN-LMEF) is proposed. Initially, GAN-LMEF uses nondominated solutions to construct a series of approximate piecewise manifolds. Then, by adopting the proposed GAN-based interpolation method, various solutions are synthesized along manifolds to fill gaps between the manifolds. Next, the entire manifold is updated from the generative population, and more promising solutions are reinterpolated along the new manifolds. The above procedures are repeated until the termination conditions are met. The contributions of this work are summarized as follows.

\begin{enumerate}

\item The proposed framework uses manifold knowledge and utilizes a GAN as a powerful weapon to learn the characteristics of solutions and synthesis solutions along the manifold. These generative solutions are of high quality and can be helpful in the evolutionary process. This provides a novel way to solve LSMOPs.

\item Considering that the gaps on the manifold vary with the evolutionary process, we propose three different interpolation strategies to better fill the gaps and maintain the manifold.

\item In the selection process, we propose a manifold selection mechanism to predict whether a generative solution is excellent without evaluating objective functions, which can reduce evaluations and avoid many meaningless evaluations.




\end{enumerate}

The rest of this paper is organized as follows. Section II introduces existing LSMOAs for LSMOPs, and GANs are briefly reviewed. Section III details the proposed algorithm for GAN-LMEF. In section IV, the empirical results of GAN-LMEF and various state-of-the-art LSMOAs on LSMOPs are presented. Finally, conclusions are drawn in Section V.

\begin{figure*}[htbp]
  \centering   
  \includegraphics[width=17.4cm]{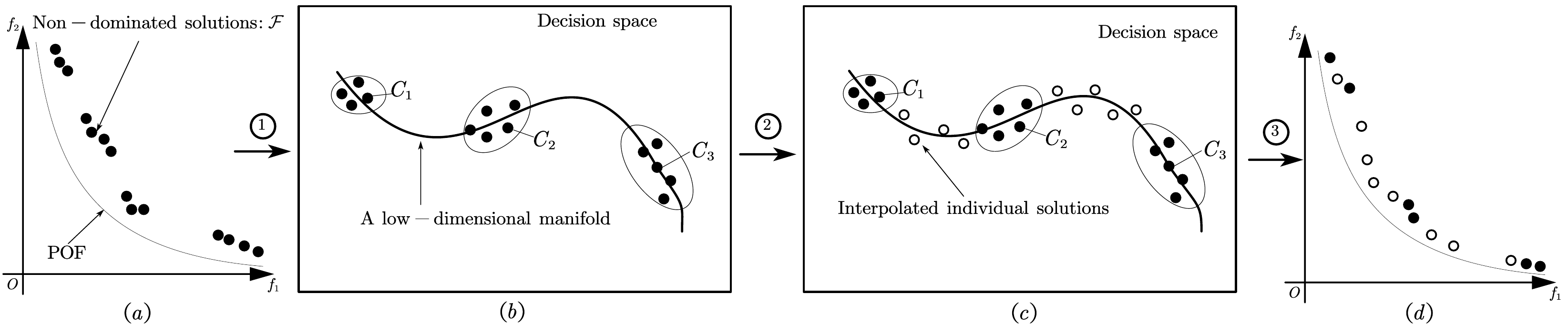}
  \caption{Step 1. Procedure {\sc Computing Central Solutions}: Map nondominated solutions $\mathcal{F}$ into an $m-1$-dimensional manifold, cluster them into several clusters, and identify the central solutions $C_i$ for each cluster. Step 2. Procedure {\sc Manifold Interpolation}: A GAN is utilized to interpolate solutions between the central solutions $C_i$. Step 3. Procedure {\sc Section}: Select the excellent solutions from the original population and interpolated solutions to form a new population to guide future searches.}
  \label{fig:tuli}
\end{figure*}

\section{PRELIMINARY STUDIES}

\subsection{Large-scale Multiobjective Optimization}
The mathematical form of LSMOPs is as follows:
\begin{equation}
\left\{
\begin{array}{lr}
\min \ F(x)=<f_1(x),f_2(x),...,f_m(x)>  & \\
s.t.\ x \in \Omega .
\end{array}
\right.
\end{equation}
where $x=(x_1,x_2,...,x_n)$ is the $n$-dimensional decision vector, and $F=(f_1,f_2,...,f_m)$ is the $m$-dimensional objective vector. It should be noted that the dimension of the decision vector $n$ is greater than 100\cite{8681243}. The goal of large-scale multiobjective optimization algorithms is to find the solutions for which all objectives are as small as possible. However, conflicts exist between multiple bjectives, and one solution cannot satisfy the minimum of all objectives. Therefore, a trade-off method called Pareto dominance is introduced to compare these solutions. The set of optimal trade-off solutions is called the POS in the decision space and the Pareto-optimal front (POF) in the objective space.
\begin{myDef}
	\textbf{(Decision Vector Domination)} A decision vector $x_1$ Pareto-dominates another vector $x_2$ denoted by $x_1\succ x_2$, if and only if
	\begin{equation}\label{chap2:equ2}
	\begin{cases}
	f_i(x_1)& \leq f_i(x_2),\quad \forall i=1,...,m\\
	f_i(x_1)& < f_i(x_2),\quad \exists i=1,...,m.
	\end{cases}
	\end{equation}
	
\end{myDef}

\begin{myDef}
	\textbf{(Pareto-Optimal Set, POS)} If a decision vector $x^*$  satisfies
	\begin{equation}
	POS=\left\{x^*|\nexists x, x\succ_t x^*\right\},
	\end{equation}
	then all $x^*$ are called Pareto-optimal solutions, and the set of Pareto-optimal solutions is called the POS.
\end{myDef}

\begin{myDef}
	\textbf{(Pareto-Optimal Front, POF)} POF is the corresponding objective vector of the POS
	\begin{equation*}
	POF=\{y^*|y^*=F(x^*),x^*\in POS\}.
	\end{equation*}
\end{myDef}

\subsection{Theory of the Manifold Structure}

The Karush–Kuhn–Tucker condition induces that the POS of a continuous multiobjective problem to follow a continuous $(m-1)$-dimensional manifold in the decision space and the theorem is given in detail in \cite{Hillermeier2001Nonlinear,Mardle1999Nonlinear}. Suppose that the objectives $f_i(x)$, $i=1,...m$ are continuous and differentiable at the Pareto-optimal solution $x^*$, then $\alpha=(\alpha_1,...\alpha_m)^T$ ($||\alpha||_2=1$) exists and satisfies
\begin{equation}\label{eq:kkt}
\sum_{i=1}^m \alpha_i \nabla f_i(x^*)=0
\end{equation}
Points satisfying Equation (\ref{eq:kkt}) are Karush–Kuhn–Tucker points. Equation (\ref{eq:kkt}) has $n+1$ equality constraints and $n+m$ variables $x_1,..x_n,\alpha_1,...,\alpha_m$. Thus, under certain smoothness conditions, the distribution of POS of a multiobjective problem is a continuous ($m-1$)-dimensional manifold. Test instances in the LSMOP benchmark\cite{Cheng2017Test} are continuous and differentiable and meet this basic theorem.

\subsection{Related Work}

Existing approaches for large-scale multiobjective optimization can be roughly classified into three different categories as follows.

The first category is the decision variable analysis-based approaches. Zhang \textit{et al.} \cite{7544478} proposed a decision variable clustering-based large-scale evolutionary algorithm (LMEA). In LMEA, the decision variables are divided into two types using a clustering method. Then, for diversity-related variables and convergence-related variables, two different strategies are implemented to improve convergence and diversity.
 
Ma \textit{et al.}\cite{7155533} presented a multiobjective evolutionary algorithm based on decision variable analysis (MOEA/DVA). The LSMOP is decomposed into a number of simpler subproblems that are regarded as an independent subcomponent to optimize. The disadvantage of decision variable analysis-based LSMOAs is that they may incorrectly identify the linkages between decision variables to update solutions, which may lead to local optima \cite{7544478}.



The second category applies the decision variable grouping framework \cite{Yang2014Large}. Antonio \textit{et al.} \cite{6557903} maintained several independent subpopulations. Each subpopulation is a subset of the equal-length decision variables obtained by variable grouping (e.g., random grouping, linear grouping, ordered grouping, or differential grouping). All of the subpopulations work cooperatively to optimize the LSMOPs in a divide-and-conquer manner.

Tian \textit{et al.} \cite{8681243} used a competitive swarm optimizer to solve LSMOPs. The proposed LMOCSO suggests a two-stage strategy to update the positions of particles: a competitive mechanism is adopted to determine the particles to be updated, and the proposed updating strategy is used to update each particle. However, for decision variable grouping-based LSMOAs, two related variables may be divided into different groups, which may lead to local optima \cite{Yang2014Large}. 

Shang \textit{et al.} \cite{s7091892} presented the idea of cooperative coevolution, which is adopted to address large-scale multiobjective capacitated routing problems. Bergh \textit{et al.} \cite{Bergh2004A} proposed a method of casting particle swarm optimization into a cooperative framework. It partitions the search space into lower-dimensional subspaces to overcome the exponential increase in difficulty and guarantees that it will be able to search every possible region of the search space.

The third category is based on problem transformation. He \textit{et al.}\cite{8628263} introduced a generic framework called the large-scale multiobjective optimization framework (LSMOF). LSMOF reformulates the original LSMOP into a low-dimensional single-objective optimization problem with some direction vectors and weight variables, aimed at guiding the population towards the POS.

Zille \textit{et al.} \cite{zille2017framework} proposed a weighted optimization framework (WOF) for solving LSMOPs. Decision variables are divided into many groups, and each group is assigned a weight variable. Then, in the same group, the weight variables are regarded as a subproblem of a subspace of the original decision space.

All of these categories heavily depend on genetic operators (crossover and mutation) to reproduce offspring. Recently, He \textit{et al.}\cite{He2020Evolutionary} designed a GAN-driven evolutionary multiobjective optimization (GMOEA) that generates offspring by GANs. However, as previously mentioned, these algorithms ignore the fact that Pareto-optimal solutions are uniformly distributed on the manifold in the low-dimensional space. Therefore, in this work, we focus on how to employ GANs to fix and maintain the manifold of solutions to better solve LSMOPs.

  \begin{figure*}[htbp]
  \centering   
  \includegraphics[width=17.6cm]{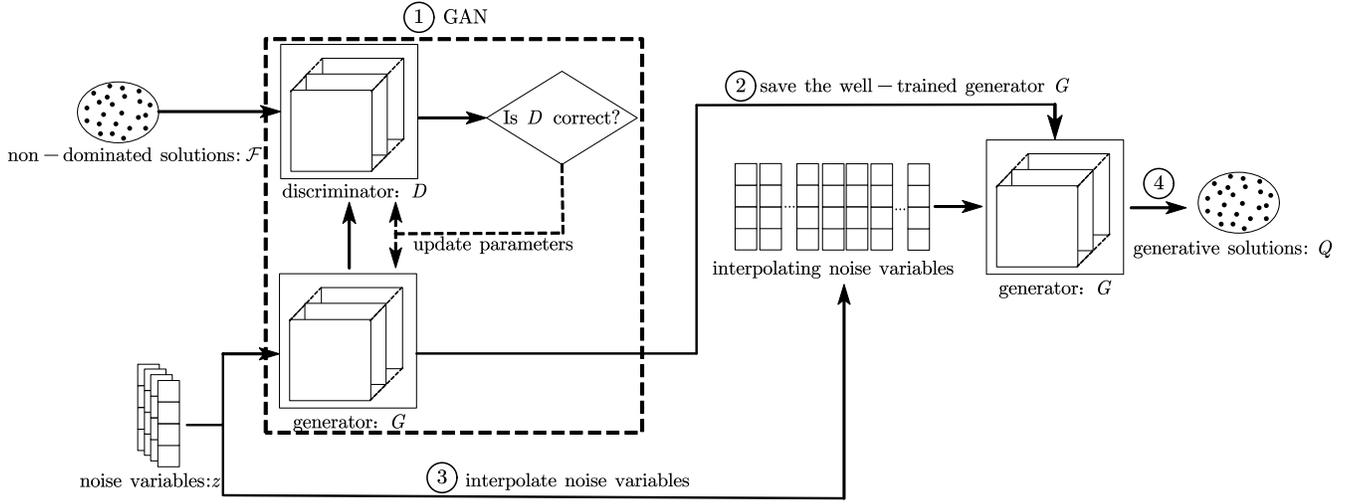}
  \caption{The flowchart of the manifold interpolation model. Step 1. Train the GAN with nondominated solutions $\mathcal{F}$ and noise variables $z$. Step 2. Save the well-trained $generator: G$. Step 3. Apply interpolation strategies to $z$ and input these interpolating noise variables into the $generator: G$. Step 4. Generate interpolating solutions $Q$. }
  \label{fig:gan}
\end{figure*}

\subsection{Generative adversarial networks}

GANs are generative models used to learn the characteristics of complicated real-world data. The basic structure of a GAN contains two neural nets: a discriminator and a generator. In addition to using generator neural nets to synthesize meaningful instances from priori distributions, a discriminator is trained to distinguish fake samples synthesized by the generator and real samples from the training dataset. The generator aims to learn the distribution of training instances and generate fake instances to deceive the discriminator. The discriminator consists of a classifier that can determine whether the input sample is a generative fake sample or a real sample. The training procedure culminates with a balance between the generator and discriminator, i.e., the discriminator cannot make a better decision that determines whether a particular sample is fake or real. Compared with other generative models, GANs provide a novel and efficient framework for learning and generating data. Therefore, GANs have recently been successfully applied to many applications \cite{dong2018musegan,ccfinproceedings,NIPS2017_7159,He2020Evolutionary}.


 
To learn the distribution of the training data, the generator $G$ takes the prior distribution of a random noise variable $z\sim P_z$ as input to generate the data distribution $P_G(z)$ approximating to the real data distribution $P_{data}(x)$, and tries to make the difference between distributions of the generative data and true data as small as possible. In existing GANs, Kullback–Leibler divergence\cite{do2002wavelet}, Jensen-Shannon divergence\cite{lin1991divergence}, and Wassertein distance\cite{NIPS2017_7159} can be used to measure the difference between two probability distributions. The discriminator determines whether the input is generative fake instances or true instances, and the output probability of discriminator $D_G(z)$ tends to 0 for fake instances, while tends to 1 for true instances. Finally, the discriminator cannot distinguish whether the data belong to real samples or generative samples. Mini-batch stochastic gradient descent\cite{zhang2004solving} training is used to generate the antagonistic network. The discriminator is updated by the stochastic gradient ascending method, and the generator is updated by the stochastic gradient descent method. The process of training is a game between the two neural networks, and the model eventually tends to the global optimum. The objective function of the game can be expressed as
   
\begin{equation}\label{eq:aei1}
\begin{split}
  \min_{G}\max_{D} V(G,D)=E_{x\sim P_{data(x)}}[log D(x)]\\+E_{z\sim P_{z(z)}}[log(1-D(G(z)))].
  \end{split}
\end{equation}
$G(z)$ is the data generated from the input noise variable $z$, and $x$ is the data from true instances. $x\sim P_(x)$ means that $x$ obeys the distribution of $P$.

\section{PROPOSED FRAMEWORK}
  
This work proposes a GAN-driven manifold interpolation method to address LSMOPs. 
The details of the proposed algorithm, GAN-LMEF, are presented in this section. 
Briefly, GAN-LMEF consists of three subroutines: Procedure {\sc Computing Central Solutions}, Procedure {\sc Manifold Interpolation}, and Procedure {\sc Selection}. In Fig. \ref{fig:tuli}, we give a detailed illustration of the main loop in GAN-LMEF. 
The main purpose of Procedure {\sc Computing Central Solutions} is to map solutions into a manifold and find some representative solutions, called central solutions, in this manifold. 
In Procedure {\sc Manifold Interpolation}, a GAN is utilized to interpolate solutions between central solutions. 
Procedure {\sc Selection} selects excellent solutions from the original population and generative solutions to form a new population to guide future searches.
In the following subsections, we will introduce the three main components: Procedure {\sc Computing Central Solutions}, Procedure {\sc Manifold Interpolation}, Procedure {\sc Selection}.

\subsection{Computing Central Solutions procedure}

Central solutions are the representatives on the manifold, and interpolating between these central solutions may yield better solutions. First, to obtain the manifold, the nondominated solutions $\mathcal{F}$ are mapped as the manifold data $\phi(\mathcal{F})$ by principal component analysis (PCA). Then, the $k$-means algorithm \cite{10.2307/2346830} is performed to cluster solutions with similar manifold characteristics. The mapped data $\phi(\mathcal{F})$ are clustered into $k$ clusters. Each cluster is denoted as $\phi(\mathcal{F})_i$, $i=1,...,k$, and in each cluster, a central solution is identified to represent the cluster. The central solution $C_i$ for each cluster $\phi(\mathcal{F})_i$ is defined as the solution with the minimum distance from $\phi(C_i)$ to the centroid of the cluster be calculated as
\begin{equation}\label{eq:cen}
  C_i=\mathop{\arg\min}_{x} || \phi(x)- \overline{\phi(\mathcal{F})_i}   ||,
 \end{equation}
 The centroid $\overline{\phi(\mathcal{F})_i} $ of the cluster $\phi(\mathcal{F})_i$ is
 \begin{equation}\label{eq:cen2}
 \overline{\phi(\mathcal{F})_i}=\frac{\sum_{j=1}^{|\phi(\mathcal{F})_i|} \phi(\mathcal{F})_{i,j} }{|\phi(\mathcal{F})_i|},
 \end{equation}
where $\phi(\mathcal{F})_{i,j}$ represents the $j$-th data in the cluster $\phi(\mathcal{F})_i$.

The central solutions identified and the piecewise manifolds are shown in Fig. \ref{fig:tuli} (b). The details of computing the central solutions are given in Procedure {\sc Computing Central Solutions}.

\begin{procedure}
  \caption{(\sc Computing Central Solutions)}
  \label{alg:getcen}
  \KwIn{$\mathcal{F}$ (the non-dominated set), $k$ (the number of clusters)}
  \KwOut{$C$ (the central solutions)}
  $C=\varnothing$\;
  Use PCA with $\mathcal{F}$ to get the manifold $\phi(\mathcal{F})$\;
  Cluster $\phi(\mathcal{F})$ into $k$ clusters by using $k$-means\;
  \For{$i=1,...,k$}{
  Calculate central solution $C_{i}$ for each cluster $\phi(\mathcal{F})_{i}$ according to Equation (\ref{eq:cen})\;
  $C=C\cup{C_{i}}$\;
  }
  \Return $C$
\end{procedure}

\subsection{Manifold Interpolation procedure}

\begin{procedure}
  \caption{(\sc GAN Manifold Interpolation)}
  \label{alg:mi}
  \KwIn{$C$ (the central solutions), $P$ (the population), $k$ (the number of clusters)}
  \KwOut{$Q$ (the interpolation solutions)}
  $Q=\varnothing$\;
  $\mathcal{F}=\textbf{Fast-Non-Dominated-Sort}(P)$\;
  Sample noise variables $\{z\}$ from noise prior $U(-1,1)$\;
  Train a GAN with $\mathcal{F}$ and $\{z\}$, save its $generator$: $G$\;
  \For{$i=1,...,k$}{
  Extract the corresponding variable $z_{C_i}$ from $\{z\}$ where $G(z_{C_i})=\hat{C_i}$\;
  }
  //Interpolating between clusters\;
  \For{$i=1,...,k$}{
  	\For{$j=1,...,k$ and $j \neq i$}{
  	Interpolate the noise variable $z_{ij}$ according to Equation (\ref{eq:aei1})\;
  	$Q=Q \cup G(z_{ij})$\;
  	}
  }
  //Interpolating in the cluster\;
  \For{$i=1,...,k$}{
  	\For{$x_a$,$x_b$ in $k$-th cluster}{
  	Interpolate the noise variable $z_{ab}$ according to Equation (\ref{eq:aei2})\;
  	$Q=Q \cup G(z_{ab})$\;
  	}
  }
  //Perturbation interpolation\;
  Add disturbance to noise variables $\{z\}$ to generate variable $\{z^'\}$\;
  $Q=Q \cup G(\{z^'\})$\;
  \Return $Q$
\end{procedure}

The manifold interpolation operations will produce a number of nonexistent solutions that are distributed on the manifold and fill gaps of this manifold. These solutions are promising and beneficial to the evolutionary process.

Before interpolating, the GAN is trained with nondominated solutions $\mathcal{F}$ and noise variables $\{z\}\sim U(-1,1)$ to learn the manifold characteristics of solutions. After that, the well-trained $generator: G$ and noise variables $\{z\}$ are saved for interpolation. In our work, three interpolation strategies are used: 1) interpolating between clusters, 2) interpolating in the cluster, and 3) perturbation interpolation.



1) Interpolating between clusters: This method is used to fill the gap between two clusters. Noise variables $z_{C_i}$ are extracted, where $z_{C_i}$ is the noise variable that generates the generative fake central solution $\hat{C_i}$ ($G(z_{C_i})=\hat{C_i}$). For every two noise variables $z_{C_i}$ and $z_{C_j}$ of fake central solutions, interpolating noise variables between them can be described as \cite{8627945}
\begin{equation}\label{eq:aei1}
  z_{ij}=(1-\alpha)z_{C_i}+\alpha z_{C_j},i,j=1,...,k, i \neq j
 \end{equation}
where $\alpha=l/n, l=0,1,..,n$, and $n$ represents the dimension of the decision variable. Because $z_{ij}$ are gradual between $z_{C_i}$ and $z_{C_j}$, the generative solutions $G(z_{ij})$ generated by $z_{ij}$ will also be gradual between the two clusters. For $k$ clusters, the GAN will interpolate solutions between different central solutions, so it will interpolate $k*(k-1)/2$ times.


2) Interpolating in the cluster: This method is used to fill gaps between individuals from the same cluster. The interpolating noise variables are described as
\begin{equation}\label{eq:aei2}
  z_{ab}=\frac{z_{a}+z_{b}}{2},
 \end{equation}
where $z_{a}$ and $z_{b}$ are the noise variables that generate fake solutions $x_{a}$ and $x_{b}$, respectively, and $x_{a}$ and $x_{b}$ are from the same cluster. Similarly, these noise variables $z_{ab}$ are inputted into the $generator: G$. This interpolation method imitates the semantic operation for noise variables \cite{UnsupervisedRepresentation}.

3) Perturbation interpolation: This method is used to apply disturbance on a noise variable to explore the neighbor space of a corresponding solution. Perturbation interpolation adds Gaussian noise to a random dimension of a noise variable. These noise variables are also inputted into the $generator: G$.

The details of the manifold interpolation are given in Procedure {\sc Manifold Interpolation} and an illustration of the manifold interpolation is shown in Fig. \ref{fig:gan}.

\subsection{Procedure of Selection}

In the selection procedure, high-quality solutions are identified from interpolated solutions set $Q$ and treated as the parental population $P$ for the next loop. 

\begin{figure}[htbp]
  \centering   
  \includegraphics[width=5cm]{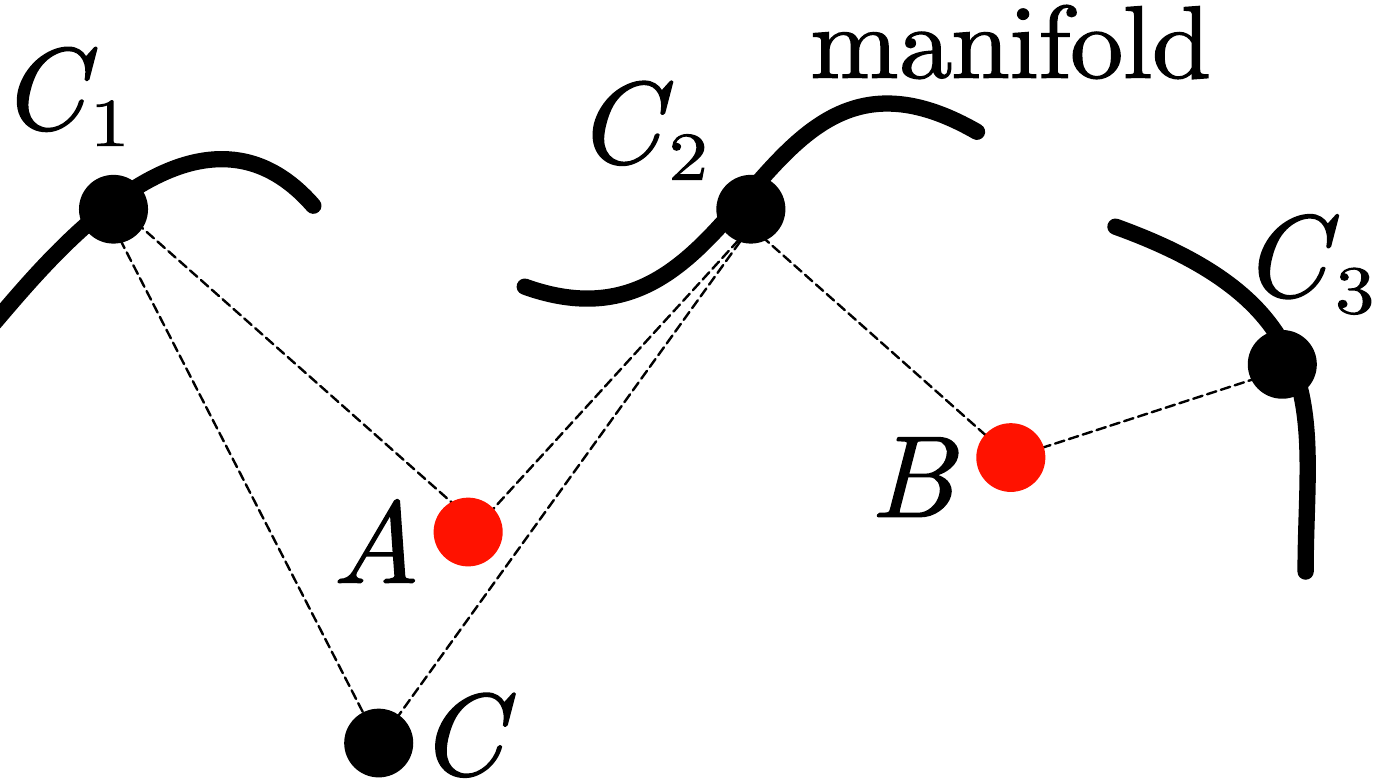}
  \caption{Manifold selection mechanism: For solutions $A$ and $C$, $C_1$ and $C_2$ are the two nearest central solutions, and for $B$, $C_2$ and $C_3$ are the two nearest central solutions. If two solutions should be selected, $A$ and $B$ are selected due to having the two minimum manifold distances.}
  \label{fig:dis2MAN}
\end{figure}

There is a huge number of generative solutions in $Q$, and it is wasteful to evaluate so many solutions to identify whether a solution is high quality. To settle this issue, a manifold selection mechanism is proposed to select solutions that have the minimum distance to the manifold, as shown in Fig. \ref{fig:dis2MAN}. 

First, PCA is employed with $C$ and $Q$ to obtain the mapping manifold data $\phi(C)$ and $\phi(Q)$, respectively.
Next, the manifold distance ($MD$) from each mapping solution $x\in\phi(Q)$ to the two nearest mapping central solutions $\phi(C_i)$ and $\phi(C_j)$ ($\phi(C_i), \phi(C_j)\in \phi(C)$) are computed 
\begin{equation}\label{eq:aei1}
  MD(x)=|||x-\phi(C_i)||_2+||x-\phi(C_j)||_2.
 \end{equation}
Then, top-($\sigma\cdot|P|$) solutions that have the minimum $MD$ are selected from $Q$ and denoted as $Q^{'}$. 
Solutions in $Q^{'}$ are promising and can improve convergence or diversity because they are close to the existing piecewise manifolds or lie in the gap between piecewise manifolds.
Finally, only solutions in $Q^{'}$ are evaluated and solutions $x_i\in P$ are replaced with $x_j\in Q^{'}$, where $x_j \succ x_i$. The new population $P$ is regarded as a parental population for the next round. 
\begin{figure}[htbp]
  \centering   
  \includegraphics[width=7.3cm]{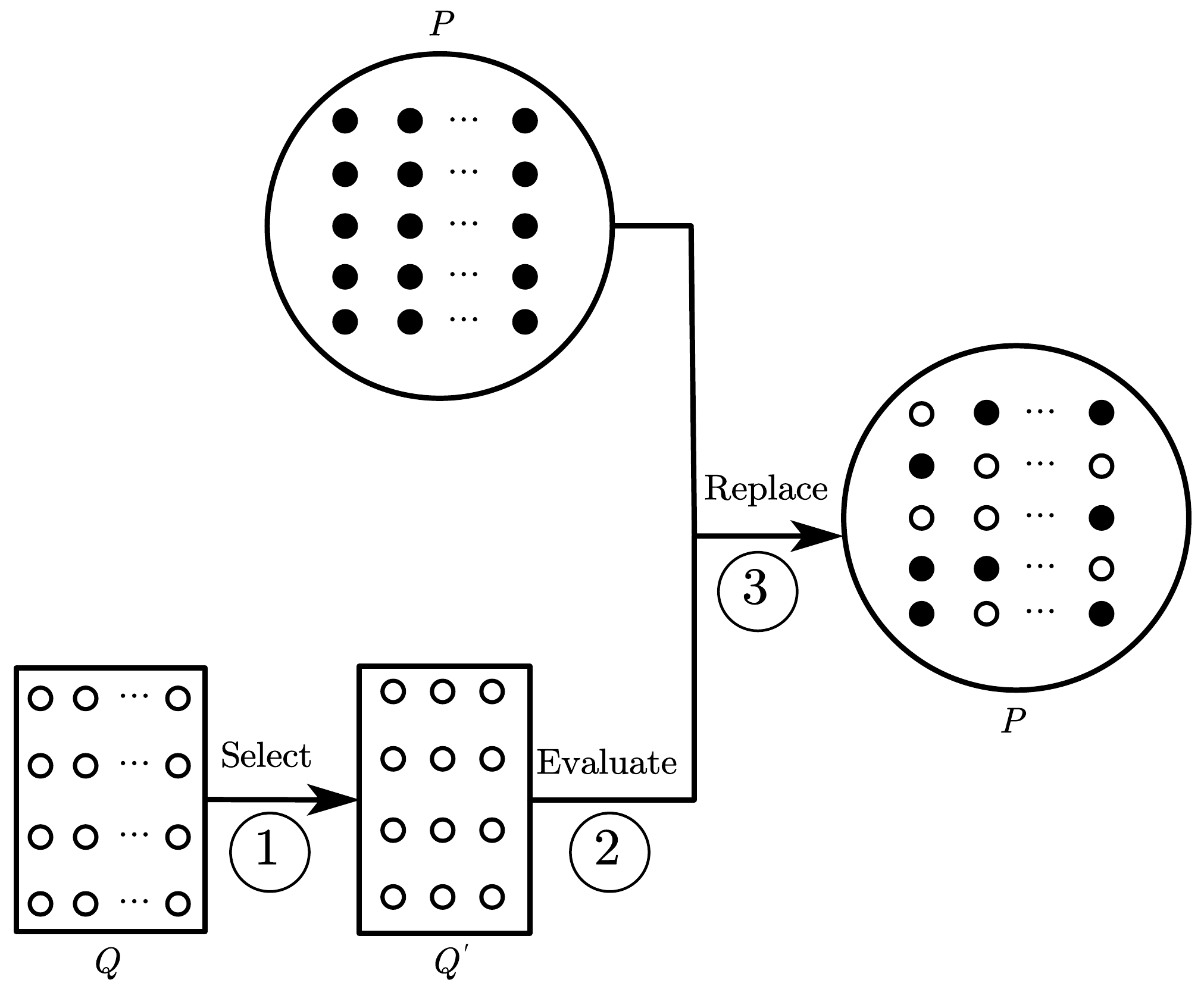}
  \caption{Step 1. Select solutions from $Q$ as $Q^{'}$ by the manifold selection mechanism. Step 2. Evaluate the solutions in $Q^{'}$. Step 3. Replace solutions in $P$ with those in $Q^{'}$.}
  \label{fig:select}
\end{figure}

Fig. \ref{fig:select} illustrates the selection process. The details of the selection are given in Procedure {\sc Selection}.

\begin{procedure}
  \caption{(\sc Selection)}
  \label{alg:select}
  \KwIn{$C$ (the central solutions), $Q$ (the interpolated solutions), $\sigma$ (the proportion of introducing interpolated solutions)}
  \KwOut{$P$ (the population)}
  Use PCA with $C$ and $Q$ to get $\phi(C)$ and $\phi(Q)$, respectively\;
  \For{$x\in \phi(Q)$}{
  	\For{$\phi(C_i), \phi(C_j)\in \phi(C)$}{
  		Find the two nearest $\phi(C_i)$ and $\phi(C_j)$ to $x$\;
  		$MD(x)=|||x-\phi(C_i)||_2+||x-\phi(C_j)||_2$\;
  	}
  }
  Select top-($\sigma\cdot|P|$) minimum $MD$ solutions from $Q$ as $Q^{'}$\;
  Evaluate solutions in $Q^{'}$\;
  Replace solutions $x_i\in P$ with $x_j\in Q^{'}$ where $x_j \succ x_i$\;
  \Return $P$
\end{procedure}

\subsection{Framework of the Proposed GAN-LMEF}

The main scheme of the proposed GAN-LMEF is presented in Algorithm \ref{alg:lmea}. First, a random population $P$ is initialized. We allocate $\epsilon\cdot e$ evaluations for the interpolation of solutions, where parameter $\epsilon$ controls the number of iterations of interpolation and $e$ is the maximum number of evaluations for solving the problem.

In every iteration of the main loop, through mating selection and mutation, a new population $P$ is generated. Then, the fast nondominated sort\cite{996017} is employed to distinguish the nondominated solution set $\mathcal{F}$. Next, the nondominated solution set $\mathcal{F}$ is inputted into Procedure {\sc Computing Central Solutions} in which these nondominated solutions are mapped into an $m-1$-dimensional manifold. These mapping solutions are clustered into several clusters, and a central solution for each cluster is identified. 
Afterwards, in Procedure {\sc Manifold Interpolation}, a GAN is utilized to interpolate the solutions $Q$ between the central solutions $C$. Finally, Procedure {\sc Selection} selects excellent solutions as a parental population $P$ for the next round. When $\epsilon\cdot e$ evaluations are used, the population $P$ is treated as the initial population $initP$, and optimized by any population-based MOA with $(1-\epsilon)\cdot e$ evaluations.




\begin{algorithm}
  \caption{GAN-LMEF}
  \label{alg:lmea}
  \KwIn{$N$ (population size), $\sigma$ (the proportion of interpolated solutions), MOA (a multi-objective optimization algorithm), $F(\cdot)$ (the LSMOP), $k$ (the number of clusters), $e$ (the maximum evaluations)}
  \KwOut{$POS$ (the POS)}
  Set parameter $\epsilon$ for controlling the number of iterations of interpolation\;
  Random initial population $P$\;
  \While{Used evaluations$\leq \epsilon\cdot e$}{
  $P^'=\textbf{Mating-Selection}(P)$\;
  $P=P\cup \textbf{Variation}(P^')$\;
  $\mathcal{F}=\textbf{Fast-Non-Dominated-Sort}(P)$\;
  $C=\textbf{Computing-Central-Solutions}(\mathcal{F},k)$\;
  $Q=\textbf{GAN-Manifold-Interpolation}(C,P,k)$\;
  $P=\textbf{Selection}(C,Q,\sigma)$\;

  }
  $initP=P$\;
  $POS= \textbf{MOA}(initP,F(\cdot),(1-\epsilon)\cdot e)$\; 
  \Return $POS$
\end{algorithm}
\subsection{Complexity Analysis}

In this subsection, we present time complexity analysis for each component of GAN-LMEF and give an overall time complexity.

For the Procedure {\sc Computing Central Solutions}, the $k$-means algorithm requires $O(k \cdot n\cdot N)$, and calculating central solutions for each cluster requires $O(N)$, where $k$ is the number of clusters and $n$ and $N$ are the dimension of the decision variables and the size of the population, respectively.

For the Procedure {\sc Manifold Interpolating}, the generator of the GAN contains a three-layer fully-connected neural network\cite{sainath2015convolutional}, and the discriminator contains a three-layer fully-connected neural network in which a sigmoid unit is utilized as the output layer. Fully connected neural networks can be regarded as special convolutional neural networks. For each convolutional layer, the time complexity is $O(n_{l} \cdot n_{l-1} \cdot M_l^2 \cdot S_l^2)$\cite{7299173}, where $n_l$ is the number of output channels in the $l$-th layer, $n_{l-1}$ is the number of input channels of the $l$-th layer, $M_l$ is the size of the output feature map, and $S_l$ is the length of the convolution kernel. The layer of a fully-connected neural network is a special convolutional layer whose $S_l$ is equal to the size of the input data $X_l$ and the size of the output feature map $M_l=1$. Therefore, it takes $O(\sum_{l=1}^d n_{l} \cdot n_{l-1} \cdot X_l^2)$ to execute the GAN once, where $d$ is the depth of the neural networks. Interpolation of solutions between central solutions of clusters requires execution of the GAN $k*(k-1)/2$ times, and the total time complexity of executing the GAN in our work is $O(k^2 \cdot( \sum_{l=1}^d n_{l} \cdot n_{l-1} \cdot X_l^2))$. The time complexity of the Procedure {\sc Manifold Interpolating} is $O(k \cdot n\cdot N)+ O(N)+O(k^2 \cdot( \sum_{l=1}^d n_{l} \cdot n_{l-1} \cdot X_l^2))=O(k^2 \cdot( \sum_{l=1}^d n_{l} \cdot n_{l-1} \cdot X_l^2))$.

The time complexity of the Procedure {\sc Selection} is $O(N\cdot k)$.


Because $k$ is a user-defined constant, the overall time complexity of GAN-LMEF is simplified as $O(\sum_{l=1}^d n_{l} \cdot n_{l-1} \cdot X_l^2)$.

\section{EXPERIMENTS}


\subsection{Algorithms in Comparison and Test Problems}

The proposed GAN-LMEF is compared against several popular algorithms including WOF\cite{zille2017framework}, LSMOF\cite{8628263}, GMOEA\cite{He2020Evolutionary}, LMOCSO\cite{Tian2019Efficient}, and LMEA\cite{7544478}. The parameters of the WOF, LMEA, LSMOF, and LMOCSO algorithms are set to defaults according to \cite{8065138} or their references. For a fair comparison, the training parameters and the structure of the GAN in GMOEA are the same as those in ours. The basic solvers of the proposed GAN-LMEF, WOF\cite{zille2017framework}, and LSMOF\cite{8628263} are unified as NSGA2\cite{10.1007/3-540-45356-3_83}, MOEADDE\cite{4633340} and CMOPSO\cite{ZHANG201863}, respectively, and we rename them according to the basic solver, e.g., GAN-NSGA2, WOF-NSGA2 and LSMOF-NSGA2.

The widely used large-scale multiobjective test problem LSMOP1-LSMOP9\cite{Cheng2017Test} is employed in our experiments. For all competitive comparisons, the number of objectives is set to 2 and the number of decision variables is set from 500 to 2000. For the ablation study, the number of objectives varies from 2 to 6, and the number of decision variables is fixed to 500.

\subsection{Performance Indicators}
In this study, the following metrics are utilized to evaluate the performance of compared algorithms.

1) Inverted Generational Distance (IGD)\cite{1197687}: The IGD is a metric for quantifying the convergence of the solutions obtained by a multiobjective optimization algorithm. When the IGD value is small, the convergence of the solution is improved. IGD is defined as

  \begin{equation}\label{chap2:equ2}
      IGD(POF^{*},POF)=\frac{1}{n}\sum_{p^{*}\in POF^{*}}\min_{p\in POF}{\left \| p^*-p \right \|}^2,
  \end{equation}
where $POF^{*}$ is the true POF of a multiobjective optimization problem, $POF$ is an approximation set of POF obtained by a multiobjective optimization algorithm and $n$ is the number of solutions in the $POF^{*}$.

2) Schott’s Spacing (SP)\cite{8605765}: It measures the uniformity of the solutions found by the algorithm. The smaller the SP value is, the more uniform the distribution of the solution obtained by the algorithm is. It is calculated by
  \begin{equation}\label{equ:SP}
    SP=\sqrt{\frac{1}{n_{POF^*}-1}\sum_{i=1}^{n_{POF^*}}(E_i-\overline{E})^2},
  \end{equation}
  where $E_i$ denotes the Euclidean distance between the $i$-th solution in $POF^*$ and its closest solution and $\overline{E}$ represents the mean of $E_i$.

\subsection{Parameter Settings}

\begin{enumerate}

\item Population Size: The population size $N$ is set to 100 for all test instances.

\item Termination Condition: The number of maximum evaluations $e$ is set to 100,000 for all compared LSMOAs.

\item The structure of the GAN is as follows: The number of nodes of the three-layer fully-connected neural network in the generator is set to $n-1$, $\left \lceil n*1.5 \right \rceil$, and $n$, respectively. The number of nodes of the fully-connected neural network in the discriminator is set to $n$, $\left \lceil n*1.5 \right \rceil$, and 1, respectively. 

\item Parameters of GAN-LMEF: The number of clusters $k$ is set to 3. The proportion of the introduced interpolated solutions $\sigma$ is set to 0.4. The $\epsilon$ is set to 0.1. The learning rate of the GAN is set to 0.001, the epoch is set to 200, and the GAN is trained by full batch learning.

\end{enumerate}

\subsection{Performance on LSMOP Problems}

\begin{table*}[htbp]\scriptsize
  \centering
  \caption{MEAN AND VARIANCE VALUES OF IGD METRIC OBTAINED BY COMPARED ALGORITHMS OVER LSMOP PROBLEMS}
    \begin{tabular}{cccccccc}
    \toprule
    Problems & Dec.  & GAN-NSGA2 & WOF-NSGA2 & LSMOF-NSGA2 & GMOEA & LMOCSO & LMEA \\
    \midrule
    \multirow{3}[2]{*}{LSMOP1} & 500   & \textbf{3.23e-1(2.27e-4)} & 7.01e-1(5.87e-4)+ & 5.11e-1(4.37e-4)+ & 3.34e-1(1.62e-2)+ & 2.05e+0(1.05e+0)+ & 1.08e+1(1.45e-1)+ \\
          & 1000  & \textbf{3.22e-1(6.58e-4)} & 7.02e-1(5.61e-4)+ & 4.83e-1(3.07e-4)+ & 5.10e-1(2.52e-3)+ & 1.36e+0(1.94e-1)+ & 1.11e+1(1.42e-1)+ \\
          & 2000  & \textbf{6.45e-1(2.20e-4)} & 7.02e-1(7.06e-4)+ & 6.50e-1(3.08e-4)= & 7.00e-1(9.85e-4)+ & 2.23e+0(7.92e-2)+ & 1.18e+1(1.87e-1)+ \\
    \midrule
    \multirow{3}[2]{*}{LSMOP2} & 500   & 1.70e-2(8.71e-6) & 3.32e-2(1.49e-6)+ & 1.74e-2(4.53e-7)= & \textbf{1.64e-2(4.25e-6)=} & 4.18e-2(8.12e-5)+ & 7.30e-2(7.49e-3)+ \\
          & 1000  & \textbf{1.25e-2(6.61e-6)} & 1.90e-2(4.32e-7)+ & 1.53e-2(1.46e-7)+ & 1.38e-2(3.93e-6)+ & 2.49e-2(2.45e-5)+ & 4.05e-2(1.96e-3)+ \\
          & 2000  & \textbf{9.13e-3(5.65e-7)} & 1.26e-2(1.89e-7)+ & 9.66e-3(9.64e-8)+ & 9.56e-3(3.20e-6)+ & 1.40e-2(6.84e-5)+ & 2.27e-2(6.14e-3)+ \\
    \midrule
    \multirow{3}[2]{*}{LSMOP3} & 500   & 1.85e+0(2.51e-4) & 1.57e+0(2.99e-3)- & \textbf{1.56e+0(3.31e-3)-} & 1.57e+0(1.87e-6)+ & 1.67e+0(1.21e+1)- & 5.83e+1(5.06e+0)+ \\
          & 1000  & 2.08e+0(5.37e-3) & 1.58e+0(3.13e-3)- & \textbf{1.57e+0(2.95e-3)-} & 1.58e+0(2.25e-6)+ & 4.49e+0(7.28e+1)+ & 1.48e+3(3.10e+3)+ \\
          & 2000  & 4.13e+0(5.00e-2) & 1.58e+0(3.22e-3)- & \textbf{1.57e+0(2.96e-3)-} & 1.59e+0(7.76e-6)+ & 8.17e+0(1.23e+1)+ & 7.61e+2(6.77e+2)+ \\
    \midrule
    \multirow{3}[2]{*}{LSMOP4} & 500   & \textbf{3.62e-2(3.82e-7)} & 6.89e-2(5.93e-6)+ & 4.46e-2(2.52e-6)+ & 4.18e-2(5.15e-5)+ & 8.34e-2(1.05e-5)+ & 1.33e-1(2.13e-3)+ \\
          & 1000  & \textbf{2.81e-2(1.27e-7)} & 6.46e-2(5.87e-6)+ & 2.98e-2(8.98e-7)+ & 2.93e-2(1.71e-5)+ & 5.24e-2(2.53e-6)+ & 7.75e-2(9.36e-3)+ \\
          & 2000  & \textbf{1.54e-2(1.02e-7)} & 3.42e-2(1.41e-6)+ & 1.76e-2(2.75e-7)+ & 1.75e-2(1.00e-5)+ & 3.01e-2(9.23e-7)+ & 4.42e-2(2.64e-3)+ \\
    \midrule
    \multirow{3}[2]{*}{LSMOP5} & 500   & \textbf{7.42e-1(3.17e-4)} & 7.42e-1(3.67e-4)= & 7.42e-1(7.23e-4)= & 7.42e-1(3.34e-4)= & 1.43e+0(3.54e+0)+ & 2.30e+1(5.86e-1)+ \\
          & 1000  & \textbf{7.33e-1(6.08e-4)} & 7.41e-1(6.95e-4)+ & 7.42e-1(7.06e-4)+ & 7.42e-1(9.95e-4)= & 3.49e+0(1.58e+0)+ & 2.34e+1(7.34e-1)+ \\
          & 2000  & 7.45e-1(8.89e-4) & 7.42e-1(7.36e-4)= & \textbf{7.42e-1(6.45e-4)=} & 7.42e-1(1.19e-4)= & 5.28e+0(5.32e-1)+ & 2.47e+1(8.29e-1)+ \\
    \midrule
    \multirow{3}[2]{*}{LSMOP6} & 500   & \textbf{3.31e-1(1.36e-4)} & 4.28e-1(2.41e-4)+ & 3.48e-1(1.81e-4)+ & 6.06e-1(1.86e-3)+ & 7.84e-1(1.02e-5)+ & 2.27e+3(7.25e+3)+ \\
          & 1000  & 8.15e-1(2.57e-4) & 4.57e-1(3.00e-4)- & \textbf{3.40e-1(1.53e-4)-} & 6.74e-1(4.59e-3)- & 7.66e-1(7.10e-4)- & 4.65e+3(3.21e+3)+ \\
          & 2000  & 6.75e-1(5.01e-4) & 7.57e-1(6.02e-4)+ & \textbf{3.13e-1(1.15e-4)-} & 6.73e-1(3.03e-5)- & 7.55e-1(9.81e-4)= & 1.35e+3(2.84e+3)+ \\
    \midrule
    \multirow{3}[2]{*}{LSMOP7} & 500   & 1.82e+0(8.07e-4) & \textbf{1.51e+0(2.54e-3)-} & 1.51e+0(3.05e-3)- & 1.51e+0(5.14e-5)- & 1.41e+2(1.77e+3)+ & 7.49e+4(7.89e+3)+ \\
          & 1000  & 1.57e+0(3.95e-3) & 1.51e+0(3.00e-3)- & 1.51e+0(3.08e-3)- & \textbf{1.51e+0(1.30e-5)-} & 2.00e+3(1.03e+3)+ & 8.41e+4(1.09e+3)+ \\
          & 2000  & 3.89e+0(1.07e-1) & 1.51e+0(2.54e-3)- & 1.51e+0(3.41e-3)- & \textbf{1.51e+0(1.51e-5)-} & 5.46e+3(9.28e+3)+ & 8.88e+4(9.86e+3)+ \\
    \midrule
    \multirow{3}[2]{*}{LSMOP8} & 500   & 7.42e-1(3.17e-7) & \textbf{4.56e-1(2.48e-4)-} & 7.42e-1(6.08e-4)= & 7.42e-1(3.45e-4)= & 1.21e+0(1.45e+0)+ & 1.89e+1(3.80e-1)+ \\
          & 1000  & \textbf{6.81e-1(2.84e-4)} & 7.42e-1(6.68e-4)+ & 7.42e-1(6.33e-4)+ & 7.42e-1(1.14e-4)+ & 2.74e+0(6.94e-1)+ & 2.02e+1(5.10e-1)+ \\
          & 2000  & 7.46e-1(5.71e-4) & 7.42e-1(6.32e-4)= & 7.42e-1(6.28e-4)= & \textbf{7.42e-1(5.26e-4)=} & 3.34e+0(4.04e-1)+ & 2.07e+1(6.73e-1)+ \\
    \midrule
    \multirow{3}[2]{*}{LSMOP9} & 500   & \textbf{5.56e-1(9.51e-4)} & 8.10e-1(9.09e-4)+ & 8.07e-1(8.61e-4)+ & 6.75e-1(9.22e-4)+ & 5.99e-1(5.59e-1)+ & 5.68e+1(3.45e+0)+ \\
          & 1000  & \textbf{6.49e-1(2.06e-4)} & 8.09e-1(7.02e-4)+ & 7.97e-1(7.58e-4)+ & 6.59e-1(8.08e-3)+ & 1.08e+0(9.88e+1)+ & 5.91e+1(4.09e+0)+ \\
          & 2000  & \textbf{8.04e-1(7.73e-4)} & 8.52e-1(3.02e-4)+ & 8.09e-1(7.66e-4)+ & 8.08e-1(2.39e-2)= & 2.50e+0(1.87e+1)+ & 6.21e+1(5.20e+0)+ \\
    \midrule
    \multicolumn{3}{c}{+/=/-} & 16/3/8 & 13/6/8 & 15/7/5 & 24/1/2 & 27/0/0 \\
    \bottomrule
    \end{tabular}%
  \label{tab:igd}%
\end{table*}%

The statistical results of the average IGD and SP values over 10 runs can be found in Tables \ref{tab:igd} and \ref{tab:sp} respectively. In the tables, (+), (=) and (-) indicate that GAN-NSGA2 is statistically significantly better, indifferent, or significantly worse than the compared algorithms, respectively, by using the Wilcoxon test \cite{DERRAC20113} (0.05 significance level).

As can be observed from Table \ref{tab:igd}, it is obvious that the proposed GAN-NSGA2 exhibits better performance than the five compared algorithms for convergence. GAN-NSGA2 achieved 15 of the 27 best results, LSMOF-NSGA2 achieved 6 of the best results, GMOEA achieved 4 of the best results, and WOF-NSGA2 achieved 2 of the best results for IGD values. Specifically, GAN-NSGA2 performs better on LSMOP1, LSMOP4, and LSMOP9 under all decision variable settings but is slightly worse than LSMOF-NSGA2 on the LSMOP3 and LSMOP6 problems with 1000 and 2000 decision variables. GMOEA can obtain a set of well-converged solutions for the LMSOP7 problems with 1000 and 2000 decision variables. WOF-NSGA2 achieves better convergence over POF on LSMOP7 and LSMOP8 with 500 decision variables.

Table \ref{tab:sp} presents the SP values of the six compared algorithms. GAN-NSGA2 performs the best on 14 out of 27 test instances, followed by LMOCSO with 6 best results, LMEA with 4 best results, and WOF-NSGA2, LMOCSO, and GMOEA with 1 best result each. Specifically, GAN-NSGA2 performs better on LSMOP1, LSMOP3, LSMOP5, and LSMOP9 under all decision variable settings. In other test instances, GAN-LMEF falls slightly behind the corresponding best-performing algorithms.

The above experimental results suggest that GAN-NSGA2 can obtain a set of solutions with good convergence and diversity for most of the test problems. However, GAN-NSGA2 is worse for IGD values on LSMOP3 and LSMOP6. This may be attributed to the fitness landscapes of LSMOP3 being the adoption of Rosenbrock’s function and Rastrigin’s function and LSMOP6 mixing Rosenbrock’s function and Ackley’s function \cite{Cheng2017Test}. The mix of separable and nonseparable functions leads to a complicated fitness landscape and increases the difficulty of learning characteristics, thus degenerating the performance of the interpolated solutions.

\begin{table*}[htbp]\scriptsize
  \centering
  \caption{MEAN AND VARIANCE VALUES OF SP METRIC OBTAINED BY COMPARED ALGORITHMS OVER LSMOP PROBLEMS}
    \begin{tabular}{cccccccc}
    \toprule
    Problems & Dec.  & GAN-NSGA2 & WOF-NSGA2 & LSMOF-NSGA2 & GMOEA & LMOCSO & LMEA \\
    \midrule
    \multirow{3}[2]{*}{LSMOP1} & 500   & \textbf{6.04e-1(1.43e-1)} & 1.75e+0(3.73e-3)+ & 1.08e+0(2.00e-3)+ & 6.80e-1(1.49e-1)+ & 1.15e+0(6.47e-2)+ & 9.28e-1(9.48e-4)+ \\
          & 1000  & \textbf{5.47e-1(1.58e-1)} & 9.99e-1(1.17e-3)+ & 1.04e+0(1.62e-3)+ & 1.61e+0(1.02e-1)+ & 1.23e+0(1.23e-2)+ & 9.73e-1(1.28e-3)+ \\
          & 2000  & \textbf{7.23e-1(8.16e-1)} & 1.69e+0(3.09e-3)+ & 9.91e-1(1.19e-1)+ & 1.02e+0(1.08e-1)+ & 1.18e+0(4.12e-3)+ & 9.45e-1(1.39e-3)+ \\
    \midrule
    \multirow{3}[2]{*}{LSMOP2} & 500   & 3.31e-1(1.43e-3) & 4.65e-1(3.40e-4)+ & 3.11e-1(9.98e-5)- & 3.05e-1(1.53e-3)- & \textbf{1.17e-1(3.17e-3)-} & 7.56e-1(7.51e-4)+ \\
          & 1000  & 3.81e-1(1.46e-3) & 4.18e-1(2.29e-4)+ & 3.66e-1(1.61e-4)- & 3.00e-1(1.59e-3)- & \textbf{7.36e-2(2.69e-3)-} & 6.70e-1(6.05e-4)+ \\
          & 2000  & 3.73e-1(1.50e-3) & 4.85e-1(3.37e-4)+ & 3.85e-1(1.99e-4)+ & 3.80e-1(1.70e-3)+ & \textbf{3.97e-2(3.28e-3)-} & 7.30e-1(6.11e-4)+ \\
    \midrule
    \multirow{3}[2]{*}{LSMOP3} & 500   & \textbf{9.53e-1(4.98e-2)} & 1.00e+0(1.21e-3)+ & 1.80e+0(4.72e-3)+ & 1.02e+0(1.35e-1)+ & 1.00e+0(1.33e-3)+ & 1.28e+0(1.92e-3)+ \\
          & 1000  & \textbf{9.71e-1(2.37e-2)} & 1.00e+0(1.31e-3)+ & 1.81e+0(3.69e-3)+ & 1.05e+0(8.63e-2)+ & 2.28e+0(5.14e-1)+ & 9.93e-1(1.02e-3)+ \\
          & 2000  & \textbf{9.99e-1(1.21e-3)} & 1.00e+0(1.35e-3)= & 1.60e+0(5.21e-2)+ & 1.00e+0(1.71e-1)= & 1.00e+0(1.11e-1)= & 1.22e+0(2.32e-3)+ \\
    \midrule
    \multirow{3}[2]{*}{LSMOP4} & 500   & 3.62e-1(1.86e-3) & 6.85e-1(5.49e-4)+ & 3.03e-1(1.18e-4)- & 3.10e-1(2.55e-3)- & \textbf{2.79e-1(1.13e-3)-} & 5.42e-1(3.81e-4)+ \\
          & 1000  & 3.34e-1(1.37e-3) & 5.08e-1(3.30e-4)+ & 3.14e-1(1.30e-4)- & 3.07e-1(1.54e-3)- & \textbf{1.79e-1(1.43e-3)-} & 6.90e-1(6.29e-4)+ \\
          & 2000  & 3.84e-1(1.56e-3) & 5.44e-1(4.00e-4)+ & 3.26e-1(1.35e-4)- & 3.22e-1(1.59e-3)- & \textbf{1.09e-1(1.75e-3)-} & 7.41e-1(7.39e-4)+ \\
    \midrule
    \multirow{3}[2]{*}{LSMOP5} & 500   & \textbf{8.47e-1(1.15e-1)} & 2.08e+0(5.83e-3)+ & 1.17e+0(1.81e-3)+ & 1.01e+0(1.05e-1)+ & 8.63e-1(1.93e-2)+ & 8.80e-1(1.05e-3)+ \\
          & 1000  & \textbf{9.02e-1(2.88e-2)} & 1.44e+0(2.23e-3)+ & 1.04e+0(1.69e-3)+ & 1.01e+0(1.94e-1)+ & 1.13e+0(4.21e-3)+ & 1.01e+0(1.23e-3)+ \\
          & 2000  & \textbf{9.13e-1(1.46e-2)} & 9.16e-1(1.18e-3)= & 1.12e+0(1.56e-3)- & 1.01e+0(1.11e-1)+ & 1.24e+0(8.34e-3)+ & 9.24e-1(1.10e-3)+ \\
    \midrule
    \multirow{3}[2]{*}{LSMOP6} & 500   & 9.84e-1(9.93e-2) & \textbf{6.57e-1(5.70e-4)-} & 9.38e-1(1.00e-3)- & 1.09e+0(1.05e-2)+ & 1.00e+0(1.29e-1)+ & 1.08e+0(1.36e-3)+ \\
          & 1000  & 1.00e+0(1.48e-3) & 1.03e+0(1.65e-3)= & 9.55e-1(1.09e-3)- & 9.96e-1(1.02e-2)- & 1.00e+0(3.33e-1)= & \textbf{9.42e-1(1.02e-3)-} \\
          & 2000  & \textbf{5.18e-1(1.13e-2)} & 1.70e+0(3.88e-3)+ & 5.30e-1(3.51e-4)+ & 9.99e-1(2.44e-2)+ & 1.00e+0(1.90e-1)+ & 8.59e-1(1.00e-3)+ \\
    \midrule
    \multirow{3}[2]{*}{LSMOP7} & 500   & \textbf{9.98e-1(2.24e-2)} & 1.00e+0(1.25e-3)= & 1.81e+0(4.21e-3)+ & 2.44e+0(3.47e-1)+ & 2.25e+0(1.35e-2)+ & 1.16e+0(2.09e-3)+ \\
          & 1000  & 1.00e+0(4.20e-1) & 1.00e+0(1.55e-3)= & 2.03e+0(6.47e-3)+ & 2.40e+0(1.56e-1)+ & 8.96e-1(7.80e-3)- & \textbf{7.68e-1(6.49e-4)-} \\
          & 2000  & 2.02e+0(1.51e-2) & 4.00e+0(1.77e-2)+ & 2.04e+0(4.65e-3)= & \textbf{9.98e-1(4.51e-1)-} & 1.23e+0(1.84e-1)- & 1.18e+0(2.07e-3)- \\
    \midrule
    \multirow{3}[2]{*}{LSMOP8} & 500   & 1.25e+0(1.81e-1) & 1.00e+0(1.11e-3)- & 1.09e+0(1.57e-3)- & 1.88e+0(1.90e-1)- & 8.78e-1(4.13e-3)- & \textbf{6.71e-1(4.97e-4)-} \\
          & 1000  & 9.53e-1(2.11e-4) & 1.89e+0(5.30e-3)+ & 1.04e+0(1.29e-3)+ & 1.10e+0(1.51e-1)+ & 8.45e-1(8.62e-4)- & \textbf{8.08e-1(6.92e-4)-} \\
          & 2000  & 1.05e+0(5.17e-2) & 1.90e+0(5.16e-3)+ & \textbf{5.24e-1(3.28e-4)-} & 1.86e+0(1.63e-1)+ & 8.36e-1(1.14e-3)- & 8.21e-1(9.83e-4)- \\
    \midrule
    \multirow{3}[2]{*}{LSMOP9} & 500   & \textbf{8.83e-1(2.58e-4)} & 1.00e+0(1.14e-3)+ & 9.99e-1(1.49e-3)+ & 9.50e-1(8.61e-5)+ & 1.53e+0(3.52e-2)+ & 9.63e-1(1.43e-3)+ \\
          & 1000  & \textbf{8.26e-1(3.77e-4)} & 1.00e+0(1.24e-3)+ & 9.95e-1(1.41e-3)+ & 9.05e-1(2.81e-3)+ & 1.67e+0(7.13e-2)+ & 8.59e-1(8.29e-4)+ \\
          & 2000  & \textbf{8.27e-1(5.39e-4)} & 1.00e+0(1.35e-3)+ & 9.98e-1(1.49e-3)+ & 9.99e-1(3.70e-2)+ & 1.66e+0(4.91e-2)+ & 8.37e-1(7.56e-4)+ \\
    \midrule
    \multicolumn{3}{c}{+/=/-} & 20/5/2 & 16/1/10 & 18/1/8 & 14/2/11 & 21/0/6 \\
    \bottomrule
    \end{tabular}%
  \label{tab:sp}%
\end{table*}%

\subsection{Ablation Study}

In this subsection, we perform ablation experiments to demonstrate the effectiveness of applying the proposed framework to NSGA2, MOEADDE, and CMOPSO. In addition, we embed basic solvers into two popular frameworks for solving LSMOP, WOF, and LSMOF. Please note that NSGA2, MOEADDE, and CMOPSO are not suitable for solving LSMOPs. The statistics of IGD values can be found in Table \ref{tab:ablaIGD}.

\begin{table*}[htbp]\scriptsize
  \centering
  \caption{MEAN AND VARIANCE VALUES OF IGD METRIC OBTAINED BY COMPARED ALGORITHMS OVER LSMOP PROBLEMS }
  \resizebox{0.793\width}{!}{
    \begin{tabular}{ccccccccccc}
    \toprule
    Problems & Dec.  & GAN-NSGA2 & WOF-NSGA2 & \multicolumn{1}{c|}{LSMOF-NSGA2} & GAN-MOEADDE & WOF-MOEADDE & \multicolumn{1}{c|}{LSMOF-MOEADDE} & GAN-CMOPSO & WOF-CMOPSO & LSMOF-CMOPSO \\
    \midrule
    \multirow{3}[2]{*}{LSMOP1} & 500   & \textbf{3.23e-1(2.27e-4)} & 7.01e-1(5.87e-4)+ & \multicolumn{1}{c|}{5.11e-1(4.37e-4)+} & \textbf{1.83e-1(2.81e-2)} & 4.34e-1 (8.39e-2)+ & \multicolumn{1}{c|}{5.90e-1 (4.22e-2)+} & \textbf{2.58e-1(2.41e-2)} & 4.34e-1 (8.39e-2)+ & 5.92e-1 (4.22e-2)+  \\
          & 1000  & \textbf{3.22e-1(6.58e-4)} & 7.02e-1(5.61e-4)+ & \multicolumn{1}{c|}{4.83e-1(3.07e-4)+} & \textbf{3.35e-1(9.14e-3)} & 5.15e-1 (1.17e-1)+ & \multicolumn{1}{c|}{6.20e-1 (2.78e-2)+} & \textbf{5.11e-1 (1.17e-1)} & 5.35e-1(1.37e-4)+ & 6.23e-1 (2.78e-2)+ \\
          & 2000  & \textbf{6.45e-1(2.20e-4)} & 7.02e-1(7.06e-4)+ & \multicolumn{1}{c|}{6.50e-1(3.08e-4)=} & 5.28e-1(1.27e-3) & \textbf{4.97e-1 (8.38e-2)-} & \multicolumn{1}{c|}{6.45e-1 (1.58e-2)+} & \textbf{4.96e-1 (8.38e-2)} & 6.77e-1(2.24e-7)+ & 6.47e-1 (1.58e-2)+ \\
    \midrule
    \multirow{3}[2]{*}{LSMOP2} & 500   & 1.70e-2(8.71e-6) & 3.32e-2(1.49e-6)+ & \multicolumn{1}{c|}{1.74e-2(4.53e-7)=} & \textbf{2.42e-2(2.54e-6)} & 2.67e-2 (2.36e-3)+ & \multicolumn{1}{c|}{2.50e-2 (1.31e-3)+} & \textbf{2.42e-2 (1.31e-3)} & 2.67e-2 (2.36e-3)- & 3.18e-2(2.78e-8)+ \\
          & 1000  & \textbf{1.25e-2(6.61e-6)} & 1.90e-2(4.32e-7)+ & \multicolumn{1}{c|}{1.53e-2(1.46e-7)+} & \textbf{1.33e-2(1.30e-6)} & 1.81e-2 (5.03e-4)+ & \multicolumn{1}{c|}{1.81e-2 (4.96e-4)+} & \textbf{1.71e-2(4.21e-8)} & 1..81e-2 (5.03e-4)+ & 1.81e-2 (4.96e-4)+ \\
          & 2000  & \textbf{9.13e-3(5.65e-7)} & 1.26e-2(1.89e-7)+ & \multicolumn{1}{c|}{9.66e-3(9.64e-8)+} & \textbf{8.48e-3(3.16e-7)} & 1.23e-2 (3.14e-4)+ & \multicolumn{1}{c|}{1.30e-2 (3.66e-4)+} & \textbf{1.05e-2(2.24e-8)} & 1.27e-2 (3.14e-4)+ & 1.33e-2 (3.66e-4)+  \\
    \midrule
    \multirow{3}[1]{*}{LSMOP3} & 500   & 1.85e+0(2.51e-4) & 1.57e+0(2.99e-3)- & \multicolumn{1}{c|}{\textbf{1.56e+0(3.31e-3)-}} & 1.51e+0(2.31e-2) & \textbf{1.50e+0 (4.14e-2)=} & \multicolumn{1}{c|}{1.56e+0 (6.81e-4)+} & 1.78e+0(1.57e-2) & \textbf{1.54e+0 (4.14e-2)-} & 1.56e+0 (6.81e-4)- \\
          & 1000  & 2.08e+0(5.37e-3) & 1.58e+0(3.13e-3)- & \multicolumn{1}{c|}{\textbf{1.57e+0(2.95e-3)-}} & 1.58e+0(6.89e-3) & \textbf{1.54e+0 (7.62e-2)=} & \multicolumn{1}{c|}{1.55e+0 (5.82e-4)=} & 2.01e+0(1.96e-3) & 1.55e+0 (7.62e-2)- & \textbf{1.55e+0 (5.82e-4)-} \\
          & 2000  & 4.13e+0(5.00e-2) & 1.58e+0(3.22e-3)- & \multicolumn{1}{c|}{\textbf{1.57e+0(2.96e-3)-}} & 1.92e+0(5.05e-1) & \textbf{1.46e+0 (1.07e-1)-} & \multicolumn{1}{c|}{1.51e+0 (3.24e-4)-} & 7.54e+0(7.78e-1) & \textbf{1.46e+0 (1.07e-1)-} & 1.57e+0 (3.24e-4)- \\
    \midrule
    \multirow{3}[0]{*}{LSMOP4} & 500   & \textbf{3.62e-2(3.82e-7)} & 6.89e-2(5.93e-6)+ & \multicolumn{1}{c|}{4.46e-2(2.52e-6)+} & \textbf{4.37e-2(3.36e-5)} & 5.66e-2 (4.60e-3)+ & \multicolumn{1}{c|}{5.16e-2 (1.39e-3)+} & 5.20e-2(3.04e-5) & 5.66e-2 (4.60e-3)+ & \textbf{5.17e-2 (1.39e-3)=} \\
          & 1000  & \textbf{2.81e-2(1.27e-7)} & 6.46e-2(5.87e-6)+ & \multicolumn{1}{c|}{2.98e-2(8.98e-7)+} & \textbf{2.55e-2(9.27e-6)} & 3.64e-2 (2.63e-3)+ & \multicolumn{1}{c|}{3.29e-2 (9.26e-4)+} & 5.31e-2(8.32e-7) & 3.61e-2 (2.63e-3)- & \textbf{3.22e-2 (9.26e-4)-} \\
          & 2000  & \textbf{1.54e-2(1.02e-7)} & 3.42e-2(1.41e-6)+ & \multicolumn{1}{c|}{1.76e-2(2.75e-7)+} & \textbf{1.73e-2(5.46e-6)} & 2.70e-2 (1.97e-3)+ & \multicolumn{1}{c|}{2.36e-2 (1.28e-3)+} & 3.18e-2(2.24e-7) & 2.76e-2 (1.97e-3)- & \textbf{2.36e-2 (1.28e-3)-} \\
    \midrule
    \multirow{3}[1]{*}{LSMOP5} & 500   & \textbf{7.42e-1(3.17e-4)} & 7.42e-1(3.67e-4)= & \multicolumn{1}{c|}{7.42e-1(7.23e-4)=} & \textbf{6.41e-1(1.14e-4)} & 7.13e-1(7.12e-3)+ & \multicolumn{1}{c|}{7.42e-1 (1.19e-6)+} & \textbf{7.42e-1(1.01e-3)} & 7.44e-1 (7.12e-3)= & 7.42e-1 (1.12e-3)= \\
          & 1000  & \textbf{7.33e-1(6.08e-4)} & 7.41e-1(6.95e-4)+ & \multicolumn{1}{c|}{7.42e-1(7.06e-4)+} & \textbf{6.88e-1(1.08e-5)} & 7.47e-1 (5.74e-3)+ & \multicolumn{1}{c|}{7.42e-1 (1.14e-6)+} & \textbf{7.38e-1(4.87e-4)} & 7.47e-1 (5.74e-3)+ & 7.42e-1 (1.46e-3)= \\
          & 2000  & 7.45e-1(8.89e-4) & 7.42e-1(7.36e-4)= & \multicolumn{1}{c|}{\textbf{7.42e-1(6.45e-4)=}} & \textbf{6.72e-1(1.47e-3)} & 7.48e-2 (4.32e-3)+ & \multicolumn{1}{c|}{7.42e-1 (1.14e-6)+} & 8.03e-1(7.47e-4) & 7.49e-2 (4.32e-3)- & \textbf{7.42e-1 (5.64e-3)-} \\
    \midrule
    \multirow{3}[2]{*}{LSMOP6} & 500   & \textbf{3.31e-1(1.36e-4)} & 4.28e-1(2.41e-4)+ & \multicolumn{1}{c|}{3.48e-1(1.81e-4)+} & 3.66e-1(2.63e-4) & 5.44e-1 (1.43e-1)+ & \multicolumn{1}{c|}{\textbf{3.20e-1 (4.0Se-4)-}} & 7.65e-1(2.56e-3) & 5.44e-1 (1.43e-1)- & \textbf{3.20e-1(4.0Se-4)-} \\
          & 1000  & 8.15e-1(2.57e-4) & 4.57e-1(3.00e-4)- & \multicolumn{1}{c|}{\textbf{3.40e-1(1.53e-4)-}} & 3.84e-1(3.56e-4) & 5.69e-1 (1.30e-1)+ & \multicolumn{1}{c|}{\textbf{3.35e-1 (1.98e-4)-}} & \textbf{3.51e-1(1.43e-3)} & 5.69e-1 (1.30e-1)+ & 3.55e-1(1.98e-4)= \\
          & 2000  & 6.75e-1(5.01e-4) & 7.57e-1(6.02e-4)+ & \multicolumn{1}{c|}{\textbf{3.13e-1(1.15e-4)-}} & 5.37e-1(6.94e-3) & 5.93e-1 (1.23e-1)+ & \multicolumn{1}{c|}{\textbf{3.08e-1 (5.62e-5)-}} & \textbf{4.88e-1(5.62e-5)} & 5.93e-1 (1.23c-1)- & 5.57e-1(6.70e-3)+ \\
    \midrule
    \multirow{3}[2]{*}{LSMOP7} & 500   & 1.82e+0(8.07e-4) & \textbf{1.51e+0(2.54e-3)-} & \multicolumn{1}{c|}{1.51e+0(3.05e-3)-} & \textbf{1.50e+0(8.48e-6)} & 1.54e+0(1.14e-1)= & \multicolumn{1}{c|}{1.51e+0 (1.18e-3)=} & \textbf{1.46e+0(1.18e-1)} & 1.54e+0(1.14e-1)+ & 1.50e+0 (1.18e-3)= \\
          & 1000  & 1.57e+0(3.95e-3) & 1.51e+0(3.00e-3)- & \multicolumn{1}{c|}{1.51e+0(3.08e-3)-} & 1.51e+0(2.10e-2) & 2.43e+1 (1.02e+2)+ & \multicolumn{1}{c|}{\textbf{1.51e+0 (4.61e-4)=}} & 3.33e+0(1.04e-1) & 2.43e+1 (1.02e+2)- & \textbf{1.51e+0 (4.61e-4)-} \\
          & 2000  & 3.89e+0(1.07e-1) & 1.51e+0(2.54e-3)- & \multicolumn{1}{c|}{1.51e+0(3.41e-3)-} & 4.23e+0(9.41e+0) & \textbf{1.48e+0 (9.01e-2)-} & \multicolumn{1}{c|}{1.52e+0 (3.71e-4)-} & 7.20e+0(5.15e+0) & \textbf{1.42e+0 (9.01e-2)-} & 1.51e+0 (3.71e-4)- \\
    \midrule
    \multirow{3}[2]{*}{LSMOP8} & 500   & 7.42e-1(3.17e-7) & \textbf{4.56e-1(2.48e-4)-} & \multicolumn{1}{c|}{7.42e-1(6.08e-4)=} & \textbf{6.54e-1(3.73e-5)} & 6.56e-1 (1.23e-1)= & \multicolumn{1}{c|}{7.42e-1 (1.14e-6)+} & \textbf{4.48e-1(2.83e-3)} & 6.59e-1 (1.23e-1)+ & 7.42e-1 (1.14e-6)+ \\
          & 1000  & \textbf{6.81e-1(2.84e-4)} & 7.42e-1(6.68e-4)+ & \multicolumn{1}{c|}{7.42e-1(6.33e-4)+} & \textbf{7.07e-1(7.18e-7)} & 7.45e-1 (2.98e-3)+ & \multicolumn{1}{c|}{7.42e-1 (1.14e-6)+} & \textbf{6.57e-1(8.57e-7)} & 7.45e-1 (2.98e-3)+ & 7.42e-1 (1.14e-6)+ \\
          & 2000  & 7.46e-1(5.71e-4) & 7.42e-1(6.32e-4)= & \multicolumn{1}{c|}{7.42e-1(6.28e-4)=} & \textbf{6.63e-1(2.56e-4)} & 7.43e-1 (1.71e-3)+ & \multicolumn{1}{c|}{7.42e-1 (1.14e-6)+} & \textbf{7.39e-1(1.02e-4)} & 7.43e-1 (1.71e-3)+ & 7.42e-1 (1.14e-6)+  \\
    \midrule
    \multirow{3}[2]{*}{LSMOP9} & 500   & \textbf{5.56e-1(9.51e-4)} & 8.10e-1(9.09e-4)+ & \multicolumn{1}{c|}{8.07e-1(8.61e-4)+} & \textbf{5.84e-1(2.53e-3)} & 8.10e-1(2.30e-3)+ & \multicolumn{1}{c|}{8.09e-1 (6.05e-4)+} & \textbf{7.46e-1(4.08e-6)} & 8.10e-1(2.30e-3)+ & 8.09e-1 (6.05e-4)+ \\
          & 1000  & \textbf{6.49e-1(2.06e-4)} & 8.09e-1(7.02e-4)+ & \multicolumn{1}{c|}{7.97e-1(7.58e-4)+} & \textbf{6.26e-1(3.78e-4)} & 8.11e-1(3.02e-3)+ & \multicolumn{1}{c|}{8.01e-1 (1.05e-3)+} & \textbf{7.44e-1(1.37e-5)} & 8.13e-1(3.02e-3)+ & 8.08e-1 (1.05e-3)+ \\
          & 2000  & \textbf{8.04e-1(7.73e-4)} & 8.52e-1(3.02e-4)+ & \multicolumn{1}{c|}{8.09e-1(7.66e-4)+} & \textbf{7.13e-1(1.49e-3)} & 8.11e-1(5.16e-3)+ & \multicolumn{1}{c|}{8.09e-1 (1.53e-3)+} & \textbf{7.92e-1(1.04e-3)} & 8.11e-1(5.16e-3)+ & 8.05e-1 (1.53e-3)+ \\
    \midrule
    \multicolumn{3}{c}{+/=/-} & 16/3/8 & 13/6/8 &       & 21/4/2 & 19/3/5 &       & 15/1/11 & 13/5/9 \\
    \bottomrule
    \end{tabular}%
    }
  \label{tab:ablaIGD}%
\end{table*}%

\begin{figure*}[htbp]
  \centering   
  \includegraphics[width=18.2cm]{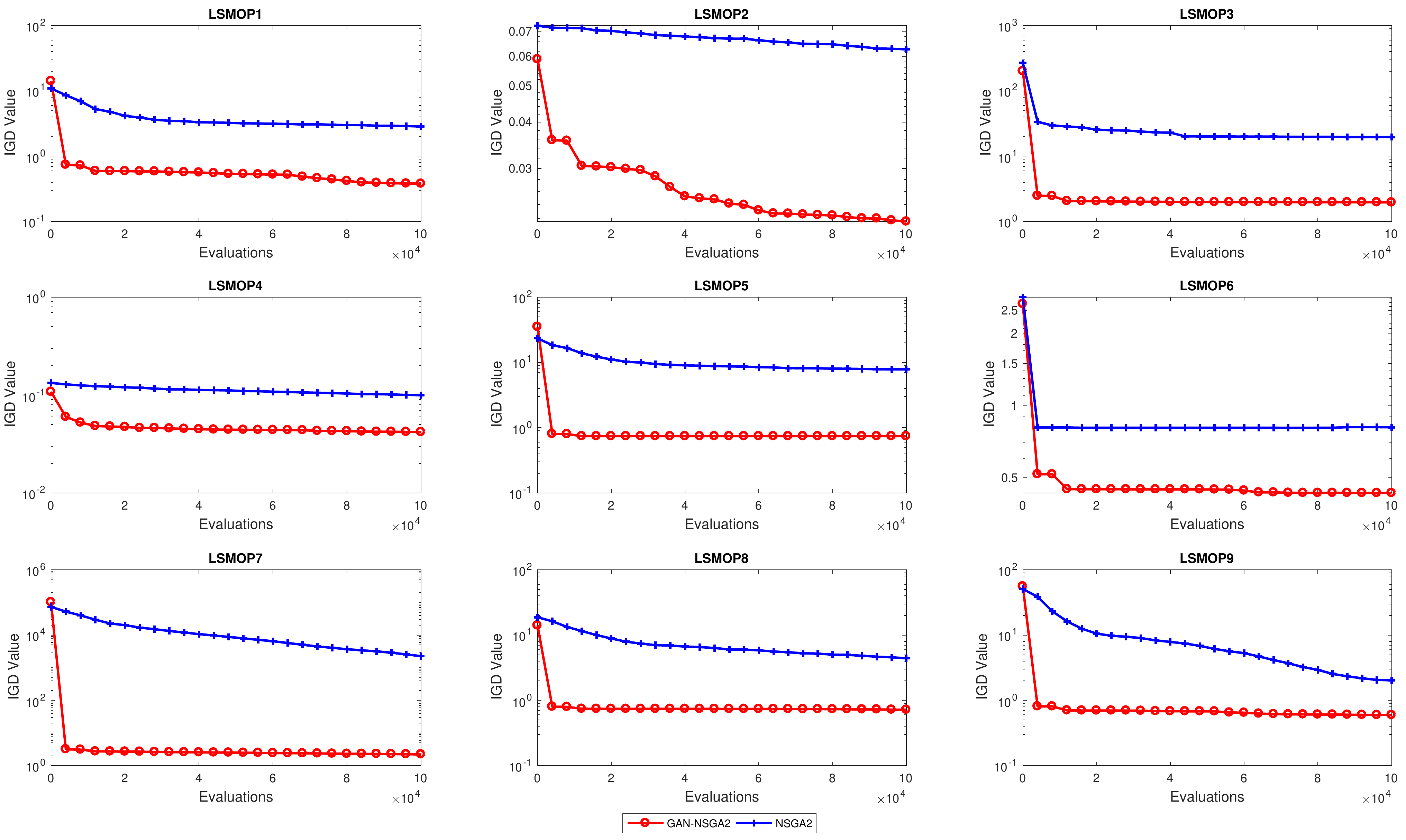}
  \caption{Convergence profiles of IGD values obtained by GAN-NSGA2 and NSGA2 on LSMOP problems with 500 decision variables.}
  \label{fig:ns}
\end{figure*}

As seen from Table \ref{tab:ablaIGD}, GAN-NSGA2 (GAN-MOEADDE and GAN-CMOPSO) achieves significantly better IGD values than WOF-NSGA2 (WOF-MOEADDE and WOF-CMOPSO) and LSMOF-NSGA2 (LSMOF-MOEADDE and LSMOF-CMOPSO). Specifically, GAN-NSGA2 performs significantly better than WOF-NSGA2 in 16 out of 27 test functions and LSMOF-NSGA2 in 13 out of 27 test functions. GAN-MOEADDE has better IGD values than WOF-MOEADDE in 21 out of 27 test functions and LSMOF-MOEADDE in 19 out of 27 instances. GAN-CMOPSO obtains better IGD values than WOF-CMOPSO in 15 out of 27 instances and LSMOF-CMOPSO in 13 out of 27 functions. These pairwise comparisons have demonstrated the ability of our proposed GAN-LEMF framework to improve the performance of existing MOAs on LSMOPs.

To verify the effectiveness of interpolation for latent noise variables $z$, we develop an algorithm called Ip-LMEF that directly interpolates solutions between the decision variables of central solutions other than latent noise variables. The piecewise linear interpolation method is used to generate solutions directly. The experimental results are shown in Table \ref{tab:ipnsga}.

\begin{table*}[htbp]\scriptsize
  \centering
  \caption{MEAN AND VARIANCE VALUES OF IGD METRIC OBTAINED BY COMPARED ALGORITHMS OVER LSMOP PROBLEMS WHEN THE NUMBER OF DECISION VARIABLES IS 500}
  \resizebox{0.833\width}{!}{
    \begin{tabular}{crcccccccc}
    \toprule
    Problems & \multicolumn{1}{c}{GAN-NSGA2} & Ip-NSGA2 & \multicolumn{1}{c|}{NSGA2} & GAN-MOEADDE & Ip-MOEADDE & \multicolumn{1}{c|}{MOEADDE} & GAN-CMOPSO & Ip-CMOPSO & CMOPSO \\
    \midrule
    LSMOP1 & \multicolumn{1}{c}{\textbf{3.23e-1(2.27e-4)}} & 6.53e-1(1.05e-3)+ & \multicolumn{1}{c|}{1.80e+0(1.57e+0)+} & \textbf{1.83e-1(2.81e-2)} & 1.05e+0(1.72e-3)+ & \multicolumn{1}{c|}{6.40e-1(9.88e-4)+} & \textbf{2.58e-1(2.41e-2)} & 6.96e-1(6.86e-4)+ & 6.55e-1(9.46e-4)+ \\
    \midrule
    LSMOP2 & \multicolumn{1}{c}{\textbf{1.70e-2(8.71e-6)}} & 4.75e-1(5.41e-4)+ & \multicolumn{1}{c|}{5.91e-2(1.40e-5)+} & \textbf{2.42e-2(2.54e-6)} & 1.19e-1(3.67e-5)+ & \multicolumn{1}{c|}{4.56e-2(4.71e-6)+} & \textbf{3.18e-2(2.78e-8)} & 1.91e-1(5.18e-5)+ & 5.55e-2(6.79e-6)+ \\
    \midrule
    LSMOP3 & \multicolumn{1}{c}{1.85e+0(2.51e-4)} & \textbf{7.45e-1(1.36e-3)-} & \multicolumn{1}{c|}{1.87e+1(7.11e+1)=} & 1.51e+0(2.31e-2) & 2.22e+0(1.19e-2)+ & \multicolumn{1}{c|}{\textbf{4.57e-2(5.36e-6)-}} & 1.78e+0(1.57e-2) & 7.43e-1(7.82e-4)- & \textbf{8.76e-2(1.97e-5)-} \\
    \midrule
    LSMOP4 & \multicolumn{1}{c}{\textbf{3.62e-2(3.82e-7)}} & 6.12e-1(8.24e-4)+ & \multicolumn{1}{c|}{9.15e-2(9.31e-5)+} & \textbf{4.37e-2(3.36e-5)} & 2.95e-1(2.09e-4)+ & \multicolumn{1}{c|}{8.98e-2(2.07e-5)+} & \textbf{6.99e-2(3.04e-5)} & 2.32e-1(7.60e-5)+ & 7.99e-2(1.44e-5)+ \\
    \midrule
    LSMOP5 & \multicolumn{1}{c}{\textbf{7.42e-1(3.17e-4)}} & 1.06e+0(2.31e-3)+ & \multicolumn{1}{c|}{5.69e+0(6.28e+0)+} & \textbf{6.41e-1(1.14e-4)} & 1.85e+0(7.04e-3)+ & \multicolumn{1}{c|}{7.31e-1(1.29e-3)+} & \textbf{7.42e-1(1.01e-3)} & 7.97e-1(5.05e-4)+ & 3.04e+0(1.97e-2)+ \\
    \midrule
    LSMOP6 & \multicolumn{1}{c}{\textbf{3.31e-1(1.36e-4)}} & 7.61e-1(1.28e-3)+ & \multicolumn{1}{c|}{8.07e-1(8.27e-3)+} & \textbf{3.66e-1(2.63e-4)} & 2.07e+0(9.43e-3)+ & \multicolumn{1}{c|}{7.31e-1(1.14e-3)+} & 7.65e-1(2.56e-3) & \textbf{4.37e-1(2.71e-4)-} & 7.92e-1(1.38e-3)+ \\
    \midrule
    LSMOP7 & \multicolumn{1}{c}{1.82e+0(8.07e-4)} & \textbf{4.81e-1(4.75e-4)-} & \multicolumn{1}{c|}{1.50e+1(1.18e+8)+} & 1.50e+0(8.48e-6) & 2.94e+0(2.07e-2)+ & \multicolumn{1}{c|}{\textbf{7.31e-1(1.18e-3)-}} & 1.46e+0(1.18e-1) & 1.00e+0(1.42e-3)+ & \textbf{7.92e-1(1.42e-3)-} \\
    \midrule
    LSMOP8 & \multicolumn{1}{c}{\textbf{7.42e-1(3.17e-3)}} & 8.64e-1(1.83e-3)+ & \multicolumn{1}{c|}{1.53e+0(8.96e+0)+} & 6.54e-1(3.73e-5) & 8.29e-1(1.36e-3)+ & \multicolumn{1}{c|}{\textbf{4.33e-1(4.12e-4)-}} & \textbf{4.48e-1(2.83e-3)} & 6.14e-1(5.35e-4)+ & 9.98e-1(2.45e-3)+ \\
    \midrule
    LSMOP9 & \multicolumn{1}{c}{\textbf{5.56e-1(9.51e-4)}} & 5.69e-1(7.80e-4)+ & \multicolumn{1}{c|}{1.23e+0(4.46e+1)+} & \textbf{5.84e-1(2.53e-3)} & 1.41e+0(3.94e-3)+ & \multicolumn{1}{c|}{6.63e-1(3.24e-4)+} & \textbf{7.46e-1(4.08e-6)} & 9.27e-1(1.22e-3)+ & 7.84e-1(1.26e-3)+ \\
    \midrule
    +/=/- &       & 7/0/2 & 8/1/0 &       & 9/0/0 & 6/0/3 &       & 7/0/2 & 7/0/2 \\
    \bottomrule
    \end{tabular}%
    }
  \label{tab:ipnsga}%
\end{table*}%

As seen from Table \ref{tab:ipnsga}, GAN-NSGA2 achieves 7 of the 9 best IGD values, while Ip-NSGA2 achieves 2 of the best results. For the optimizer MOEADDE, GAN-MOEADDE achieves 6 of the best results. For the basic solver CMOPSO, GAN-CMOPSO performs significantly better than both Ip-CMOPSO and CMOPSO in 7 out of 9 instances. We also draw convergence profiles of IGD values obtained by GAN-NSGA2 and NSGA2 in Fig. \ref{fig:ns}. The convergence rate of GAN-NSGA2 is faster than that of NSGA2. As seen in the experimental results, incorporation of GAN into NSGA2 can greatly improve the quality and convergence speed of solutions in solving LSMOPs. These results confirm the effectiveness of our proposed interpolation method. The GAN learns the mapping relationship from the latent low-dimensional manifold space to the high-dimensional solutions, and interpolation of the latent variables $z$ can generate high-quality solutions lying on the manifold.

\subsection{Performance of Different Interpolation Strategies}

 \begin{figure*}[htbp]
  \centering   
  \includegraphics[width=18.cm]{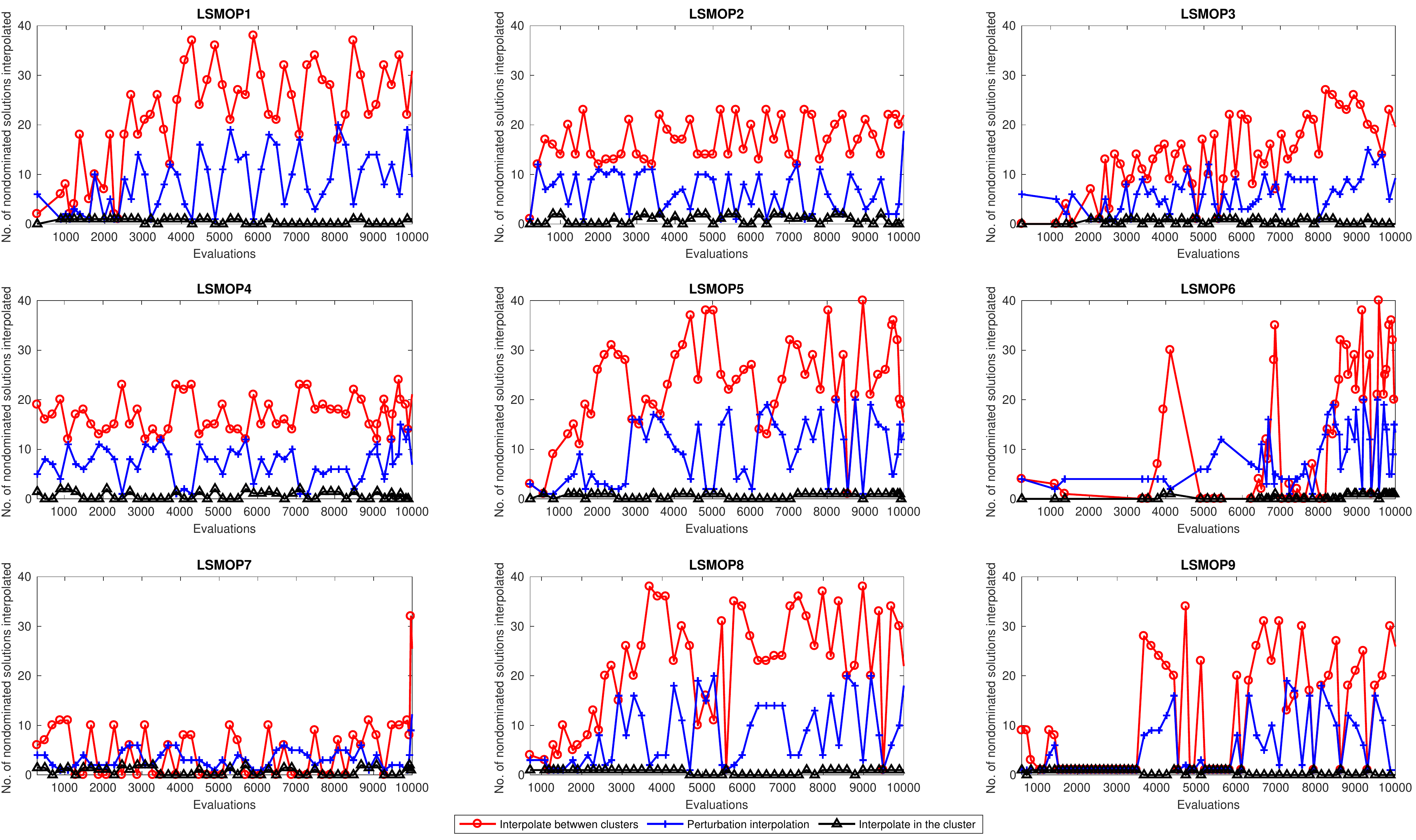}
  \caption{Number of selected nondominated solutions interpolated by different strategies on LSMOP problems with 500 decision variables.}
  \label{fig:bar}
\end{figure*}

To investigate the quality of solutions interpolated via different interpolation strategies, we record the number of different interpolated nondominated solutions selected by the Procedure {\sc Selection} during evolutionary optimization.

As shown in Fig. \ref{fig:bar}, interpolating between central solutions can produce much better individuals than the perturbation interpolation method and interpolating in the cluster method. This indicates that the gaps between clusters are always huge in solving LSMOPs, and the population lacks diversity and convergence during the entire evolutionary optimization. Therefore, interpolating between central solutions can produce more promising solutions.

\subsection{Sensitivity Study}

In GAN-LMEF, the parameters $k$ and $\epsilon$ decide the number of interpolated solutions and affect the performance. In this sensitivity study, the influence of $k$ and $\epsilon$ on the convergence of GAN-NSGA2 is investigated. Table \ref{tab:parameters} shows IGD values obtained by different choices of $k$ and $\epsilon$ values for GAN-NSGA2 on the LSMOP benchmark. When $k$ increases from 2 to 3, IGD values improve significantly in 12 out of 27 instances and decrease significantly in 7 out of 27 instances. When $\epsilon$ varies from 0.1 to 0.5, half of the instances in terms of IGD values become small, while half of the instances increase. Based on the experimental results, a larger $k$ value can generate more piecewise manifolds, thus interpolating many more solutions in the gaps. While the choice of parameter $\epsilon$ depends on the specific problem, it needs to be set carefully.


\begin{table}[htbp]\scriptsize
  \centering
  \caption{MEAN AND VARIANCE VALUES OF IGD METRIC OBTAINED BY DIFFERENT $\epsilon$ AND $k$ VALUES OF GAN-NSGA2 WHEN THE NUMBER OF DECISION VARIABLES IS 500}
    \begin{tabular}{cccc}
    \toprule
    Problems & $\epsilon$ & $k=2$   & $k=3$ \\
    \midrule
    \multirow{3}[2]{*}{LSMOP1} & 0.1   & 2.84e-1(4.13e-3) & 3.23e-1(2.27e-4) \\
          & 0.2   & 2.59e-1(1.71e-3) & 2.64e-1(2.27e-3) \\
          & 0.5   & 2.93e-1(1.81e-3) & 2.77e-1(1.65e-3) \\
    \midrule
    \multirow{3}[2]{*}{LSMOP2} & 0.1   & 2.01e-2(7.81e-6) & 1.70e-2(8.71e-6) \\
          & 0.2   & 1.96e-2(6.00e-6) & 1.59e-2(6.19e-6) \\
          & 0.5   & 1.81e-2(1.86e-5) & 1.56e-2(2.26e-5) \\
    \midrule
    \multirow{3}[2]{*}{LSMOP3} & 0.1   & 1.90e+0(1.04e-3) & 1.85e+0(2.51e-4) \\
          & 0.2   & 1.56e+0(3.45e-2) & 1.79e+0(3.72e-3) \\
          & 0.5   & 1.76e+0(1.18e-3) & 1.68e+0(4.10e-4) \\
    \midrule
    \multirow{3}[2]{*}{LSMOP4} & 0.1   & 3.85e-2(2.17e-5) & 3.62e-2(3.82e-7) \\
          & 0.2   & 3.66e-2(2.35e-5) & 3.68e-2(1.57e-5) \\
          & 0.5   & 4.51e-2(1.40e-5) & 3.18e-2(1.42e-5) \\
    \midrule
    \multirow{3}[2]{*}{LSMOP5} & 0.1   & 7.42e-1(1.63e-6) & 7.42e-1(3.17e-4) \\
          & 0.2   & 7.42e-1(5.83e-7) & 7.42e-1(2.56e-6) \\
          & 0.5   & 7.42e-1(2.80e-7) & 7.42e-1(2.76e-6) \\
    \midrule
    \multirow{3}[2]{*}{LSMOP6} & 0.1   & 5.47e-1(9.47e-6) & 3.31e-1(1.36e-4) \\
          & 0.2   & 6.09e-1(3.65e-5) & 7.80e-1(3.82e-4) \\
          & 0.5   & 7.93e-1(4.02e-4) & 7.83e-1(2.75e-4) \\
    \midrule
    \multirow{3}[2]{*}{LSMOP7} & 0.1   & 1.51e+0(4.30e-2) & 1.82e+0(8.07e-4) \\
          & 0.2   & 1.93e+0(6.78e-2) & 2.27e+0(2.68e-2) \\
          & 0.5   & 2.02e+0(2.71e-2) & 2.01e+0(1.75e-2) \\
    \midrule
    \multirow{3}[2]{*}{LSMOP8} & 0.1   & 7.42e-1(1.50e-7) & 7.42e-1(3.17e-7) \\
          & 0.2   & 7.42e-1(2.14e-7) & 7.42e-1(1.13e-7) \\
          & 0.5   & 7.42e-1(2.52e-7) & 7.42e-1(1.12e-7) \\
    \midrule
    \multirow{3}[2]{*}{LSMOP9} & 0.1   & 5.39e-1(2.15e-3) & 5.56e-1(9.51e-4) \\
          & 0.2   & 5.41e-1(3.00e-3) & 5.46e-1(8.94e-4) \\
          & 0.5   & 5.47e-1(3.34e-4) & 5.38e-1(4.35e-4) \\
    \bottomrule
    \end{tabular}%
  \label{tab:parameters}%
\end{table}%

Because POS can be represented in an $m-1$-dimensional manifold, the number of objectives $m$ can determine the complexity of the manifold, thus affecting the performance of interpolation. Therefore, we conduct the experiment concerning $m$, and the experimental results concerning the IGD metric are shown in Table \ref{tab:diffM}. GAN-NSGA2 achieves better IGD values in 12 out of 27 instances. However, we can immediately find that GAN-NSGA2 has more difficulty obtaining well-converged solutions on more objective functions because the manifold becomes more complicated and more difficult to learn.

\begin{table*}[htbp]\scriptsize
  \centering
  \caption{MEAN AND VARIANCE VALUES OF IGD METRIC OBTAINED BY COMPARED ALGORITHMS OVER LSMOP PROBLEMS WHEN THE NUMBER OF DECISION VARIABLES IS 500}
  \resizebox{1.01\width}{!}{
    \begin{tabular}{cccccccc}
    \toprule
    Problems & $m$     & GAN-NSGA2 & WOF-NSGA2 & LSMOF-NSGA2 & GMOEA & LMOCSO & LMEA \\
    \midrule
    \multirow{3}[2]{*}{LSMOP1} & 2     & \textbf{3.23e-1(2.27e-4)} & 7.01e-1(5.87e-4)+ & 5.11e-1(4.37e-4)+ & 3.34e-1(1.62e-2)+ & 2.05e+0(1.05e+0)+ & 1.08e+1(1.45e-1)+ \\
          & 4     & 8.44e-1(3.03e-5) & 8.65e-1(1.22e-4)+ & 8.77e-1(1.37e-6)+ & 8.17e-1(4.36e-4)- & 6.18e-1(3.56e-5)- & \textbf{4.35e-1(2.55e-4)-} \\
          & 6     & 9.53e-1(3.84e-3) & 9.50e-1(1.36e-3)= & 9.50e-1(1.36e-3)= & 9.57e-1(1.80e-4)= & 8.63e-1(9.21e-5)- & \textbf{4.45e-1(2.20e-4)-} \\
    \midrule
    \multirow{3}[2]{*}{LSMOP2} & 2     & 1.70e-2(8.71e-6) & 3.32e-2(1.49e-6)+ & 1.74e-2(4.53e-7)= & \textbf{1.64e-2(4.25e-6)=} & 4.18e-2(8.12e-5)+ & 7.30e-2(7.49e-3)+ \\
          & 4     & \textbf{1.47e-1(3.45e-5)} & 1.80e-1(2.10e-5)+ & 1.48e-1(4.93e-6)= & 1.48e-1(2.65e-5)= & 1.49e-1(2.86e-4)= & 5.11e-1(2.81e-4)+ \\
          & 6     & \textbf{2.46e-1(3.14e-5)} & 2.87e-1(6.33e-5)+ & 2.58e-1(1.66e-5)+ & 2.47e-1(5.55e-5)= & 2.67e-1(1.03e-4)+ & 6.31e-1(5.93e-4)+ \\
    \midrule
    \multirow{3}[2]{*}{LSMOP3} & 2     & 1.85e+0(2.51e-4) & 1.57e+0(2.99e-3)- & \textbf{1.56e+0(3.31e-3)-} & 1.57e+0(1.87e-6)+ & 1.67e+0(1.21e+1)- & 5.83e+1(5.06e+0)+ \\
          & 4     & 5.03e+0(1.86e-3) & 1.80e+0(5.42e-3)- & 1.79e+0(2.17e-3)- & 1.80e+0(1.16e-6)- & 1.15e+1(5.50e-1)+ & \textbf{7.11e-1(7.11e-4)-} \\
          & 6     & 3.54e+0(1.92e-1) & 1.85e+0(4.88e-3)- & 1.17e+0(5.42e-2)- & 1.85e+0(1.18e-6)- & 2.08e+1(4.62e-1)+ & \textbf{8.25e-1(9.69e-4)-} \\
    \midrule
    \multirow{3}[2]{*}{LSMOP4} & 2     & \textbf{3.62e-2(3.82e-7)} & 6.89e-2(5.93e-6)+ & 4.46e-2(2.52e-6)+ & 4.18e-2(5.15e-5)+ & 8.34e-2(1.05e-5)+ & 1.33e-1(2.13e-3)+ \\
          & 4     & \textbf{1.81e-1(8.52e-5)} & 2.37e-1(3.27e-6)+ & 2.15e-1(5.23e-5)+ & 2.13e-1(6.83e-5)+ & 2.11e-1(1.46e-3)+ & 5.57e-1(4.09e-4)+ \\
          & 6     & \textbf{2.93e-1(9.90e-5)} & 4.02e-1(9.15e-5)+ & 3.48e-1(2.11e-4)+ & 3.47e-1(2.89e-4)+ & 3.63e-1(3.41e-4)+ & 6.27e-1(4.49e-4)+ \\
    \midrule
    \multirow{3}[2]{*}{LSMOP5} & 2     & \textbf{7.42e-1(3.17e-4)} & 7.42e-1(3.67e-4)= & 7.42e-1(7.23e-4)= & 7.42e-1(3.34e-4)= & 1.43e+0(3.54e+0)+ & 2.30e+1(5.86e-1)+ \\
          & 4     & \textbf{4.60e-1(1.37e-4)} & 4.64e-1(1.81e-6)= & 4.80e-1(2.91e-6)+ & 4.87e-1(8.61e-3)+ & 1.84e+0(2.92e+0)+ & 9.33e-1(1.36e-3)+ \\
          & 6     & 1.15e+0(2.96e-2) & \textbf{7.40e-1(1.63e-3)-} & 9.56e-1(1.01e-5)- & 8.13e-1(6.66e-3)- & 2.14e+0(5.08e+0)+ & 1.18e+0(1.00e-3)= \\
    \midrule
    \multirow{3}[2]{*}{LSMOP6} & 2     & \textbf{3.31e-1(1.36e-4)} & 4.28e-1(2.41e-4)+ & 3.48e-1(1.81e-4)+ & 6.06e-1(1.86e-3)+ & 7.84e-1(1.02e-5)+ & 2.27e+3(7.25e+3)+ \\
          & 4     & \textbf{8.93e-1(6.97e-3)} & 9.29e-1(7.21e-7)+ & 9.45e-1(1.01e-4)+ & 9.07e-1(1.59e-3)+ & 1.14e+0(6.66e-2)+ & 9.15e-1(1.12e-3)+ \\
          & 6     & 1.21e+0(7.01e-1) & 1.39e+0(4.92e-5)+ & \textbf{1.20e+0(2.64e-4)=} & 1.55e+0(8.53e+1)+ & 1.40e+0(5.42e-1)+ & 1.20e+0(1.36e-1)= \\
    \midrule
    \multirow{3}[2]{*}{LSMOP7} & 2     & 1.82e+0(8.07e-4) & \textbf{1.51e+0(2.54e-3)-} & 1.51e+0(3.05e-3)- & 1.51e+0(5.14e-5)- & 1.41e+2(1.77e+3)+ & 7.49e+4(7.89e+3)+ \\
          & 4     & 1.24e+0(5.41e-6) & 1.23e+0(1.43e-1)= & 1.34e+0(1.09e-4)+ & \textbf{1.23e+0(1.79e-2)=} & 2.91e+2(9.60e+2)+ & 1.44e+0(2.59e-3)+ \\
          & 6     & 1.73e+0(9.91e-1) & 1.67e+0(3.73e-5)= & 1.86e+0(2.73e-4)+ & 3.40e+1(5.91e+1)+ & 1.58e+0(1.68e-1)- & \textbf{9.74e-1(1.64e-3)-} \\
    \midrule
    \multirow{3}[2]{*}{LSMOP8} & 2     & 7.42e-1(3.17e-7) & \textbf{4.56e-1(2.48e-4)-} & 7.42e-1(6.08e-4)= & 7.42e-1(3.45e-4)= & 1.21e+0(1.45e+0)+ & 1.89e+1(3.80e-1)+ \\
          & 4     & \textbf{4.36e-1(1.11e-4)} & 4.59e-1(6.46e-7)+ & 4.39e-1(1.19e-4)= & 4.53e-1(5.29e-3)+ & 2.12e+0(1.64e+0)+ & 8.11e-1(1.02e-3)- \\
          & 6     & 1.15e+0(5.80e-1) & \textbf{6.54e-1(2.13e-5)-} & 6.76e-1(1.08e-4)- & 1.15e+0(1.73e+0)= & 1.50e+0(1.74e-0)+ & 8.02e-1(9.56e-4)- \\
    \midrule
    \multirow{3}[2]{*}{LSMOP9} & 2     & \textbf{5.56e-1(9.51e-4)} & 8.10e-1(9.09e-4)+ & 8.07e-1(8.61e-4)+ & 6.75e-1(9.22e-4)+ & 5.99e-1(5.59e-1)+ & 5.68e+1(3.45e+0)+ \\
          & 4     & 9.87e-1(2.72e-2) & 1.46e+0(2.30e-8)+ & 2.25e+0(3.00e-2)+ & \textbf{5.68e-1(9.05e-2)-} & 6.95e-1(7.52e-1)- & 9.51e-1(1.00e-3)- \\
          & 6     & 1.70e+0(3.27e-1) & \textbf{7.29e-1(9.46e-5)-} & 3.71e+0(5.51e-2)+ & 8.74e-1(3.03e-1)- & 2.30e+0(7.69e-1)+ & 9.91e-1(1.53e-3)- \\
    \midrule
    \multicolumn{3}{c}{+/=/-} & 14/5/8 & 14/7/6 & 12/8/7 & 21/1/5 & 15/2/9 \\
    \bottomrule
    \end{tabular}%
    }
  \label{tab:diffM}%
\end{table*}

\section{CONCLUSION}

This paper has presented a GAN-based evolutionary search framework for solving LSMOPs, called GAN-LMEF. The aim of GAN-LMEF is to interpolate solutions on the manifold via a GAN to maintain the effective manifold. Based on the generative population, higher-quality offspring are reproduced, thereby improving evolutionary performance.

In the proposed GAN-LMEF, a GAN is employed to learn characteristics from nondominated solutions and generate a number of solutions by three different interpolation strategies. Then, a manifold selection mechanism selects promising solutions from the generative solutions for the next round. We integrated the proposed framework with NSGA2, MOEADDE, and CMOPSO to evaluate the performance. The experimental results have demonstrated that the proposed GAN-LMEF can better solve LSMOPs than several state-of-the-art LSMOAs.

This paper presents preliminary work. Several possible directions may be taken for future work. For large-scale problems with complicated search spaces, such as constraints and disconnected Pareto-optimal regions, the proposed framework still has much room for improvement. In addition, a rich body of transfer techniques such as evolutionary transfer optimization \cite{8114198,9122031,8100935,9199822} and advanced machine learning \cite{9002755} and  can inspire further innovations in solving real-world applications with large-scale decision variables.

\section*{Acknowledgment}

\bibliography{mybibtex}

\begin{thebibliography}{10}
\providecommand{\url}[1]{#1}
\csname url@samestyle\endcsname
\providecommand{\newblock}{\relax}
\providecommand{\bibinfo}[2]{#2}
\providecommand{\BIBentrySTDinterwordspacing}{\spaceskip=0pt\relax}
\providecommand{\BIBentryALTinterwordstretchfactor}{4}
\providecommand{\BIBentryALTinterwordspacing}{\spaceskip=\fontdimen2\font plus
\BIBentryALTinterwordstretchfactor\fontdimen3\font minus
  \fontdimen4\font\relax}
\providecommand{\BIBforeignlanguage}[2]{{%
\expandafter\ifx\csname l@#1\endcsname\relax
\typeout{** WARNING: IEEEtran.bst: No hyphenation pattern has been}%
\typeout{** loaded for the language `#1'. Using the pattern for}%
\typeout{** the default language instead.}%
\else
\language=\csname l@#1\endcsname
\fi
#2}}
\providecommand{\BIBdecl}{\relax}
\BIBdecl

\bibitem{6191315}
A.~{Ponsich}, A.~L. {Jaimes}, and C.~A.~C. {Coello}, ``A survey on
  multiobjective evolutionary algorithms for the solution of the portfolio
  optimization problem and other finance and economics applications,''
  \emph{IEEE Transactions on Evolutionary Computation}, vol.~17, no.~3, pp.
  321--344, June 2013.

\bibitem{Stanko2015Large}
Z.~P. Stanko, T.~Nishikawa, and S.~R. Paulinski, ``Large-scale multi-objective
  optimization for the management of seawater intrusion, santa barbara, ca,''
  in \emph{Agu Fall Meeting}, 2015.

\bibitem{9185798}
Z.~{Zhao}, M.~{Jiang}, S.~{Guo}, Z.~{Wang}, F.~{Chao}, and K.~C. {Tan},
  ``Improving deep learning based optical character recognition via neural
  architecture search,'' in \emph{2020 IEEE Congress on Evolutionary
  Computation (CEC)}, 2020, pp. 1--7.

\bibitem{208913}
J.~{Cheney}, ``The application of optimisation methods to the design of large
  scale telecommunication networks,'' in \emph{IEE Colloquium on Large-Scale
  and Hierarchical Systems}, March 1988, pp. 2/1--2/2.

\bibitem{wang2015memetic}
H.~Wang, L.~Jiao, R.~Shang, S.~He, and F.~Liu, ``A memetic optimization
  strategy based on dimension reduction in decision space,'' \emph{Evolutionary
  computation}, vol.~23, no.~1, pp. 69--100, 2015.

\bibitem{7155533}
X.~{Ma}, F.~{Liu}, Y.~{Qi}, X.~{Wang}, L.~{Li}, L.~{Jiao}, M.~{Yin}, and
  M.~{Gong}, ``A multiobjective evolutionary algorithm based on decision
  variable analyses for multiobjective optimization problems with large-scale
  variables,'' \emph{IEEE Transactions on Evolutionary Computation}, vol.~20,
  no.~2, pp. 275--298, 2016.

\bibitem{6557903}
L.~M. {Antonio} and C.~A.~C. {Coello}, ``Use of cooperative coevolution for
  solving large scale multiobjective optimization problems,'' in \emph{2013
  IEEE Congress on Evolutionary Computation}, June 2013, pp. 2758--2765.

\bibitem{hongye111}
H.~{Hong}, K.~{Ye}, M.~{Jiang}, and K.~{Tan}, ``Solving large-scale
  multi-objective optimization via probabilistic prediction model,'' in
  \emph{11th Edition of International Conference Series on Evolutionary
  Multi-Criterion Optimization}, 2021.

\bibitem{8628263}
C.~{He}, L.~{Li}, Y.~{Tian}, X.~{Zhang}, R.~{Cheng}, Y.~{Jin}, and X.~{Yao},
  ``Accelerating large-scale multi-objective optimization via problem
  reformulation,'' \emph{IEEE Transactions on Evolutionary Computation}, pp.
  1--1, 2019.

\bibitem{Hillermeier2001Nonlinear}
Hillermeier and Claus, ``Nonlinear multiobjective optimization: A generalized
  homotopy approach,'' \emph{Journal of the Operational Research Society}, vol.
  10.1007/978-3-0348-8280-4, no.~2, pp. 246--247, 2001.

\bibitem{Mardle1999Nonlinear}
S.~Mardle and K.~M. Miettinen, ``Nonlinear multiobjective optimization,''
  \emph{Journal of the Operational Research Society}, vol.~51, no.~2, p. 246,
  1999.

\bibitem{4358761}
Q.~{Zhang}, A.~{Zhou}, and Y.~{Jin}, ``Rm-meda: A regularity model-based
  multiobjective estimation of distribution algorithm,'' \emph{IEEE
  Transactions on Evolutionary Computation}, vol.~12, no.~1, pp. 41--63, Feb
  2008.

\bibitem{min2020EvolutionaryManifold}
M.~Jiang, Z.~Wang, L.~Qiu, S.~Guo, X.~Gao, and K.~C. Tan, ``A fast dynamic
  evolutionary multiobjective algorithm via manifold transfer learning,''
  \emph{IEEE Transactions on Cybernetics}, 2020.

\bibitem{9185009}
J.~{Zhang}, S.~{Li}, M.~{Jiang}, and K.~C. {Tan}, ``Learning from weakly
  labeled data based on manifold regularized sparse model,'' \emph{IEEE
  Transactions on Cybernetics}, pp. 1--14, 2020.

\bibitem{NIPS2014_5352}
D.~P. Kingma, S.~Mohamed, D.~Jimenez~Rezende, and M.~Welling, ``Semi-supervised
  learning with deep generative models,'' in \emph{Advances in Neural
  Information Processing Systems 27}, Z.~Ghahramani, M.~Welling, C.~Cortes,
  N.~D. Lawrence, and K.~Q. Weinberger, Eds.\hskip 1em plus 0.5em minus
  0.4em\relax Curran Associates, Inc., 2014, pp. 3581--3589.

\bibitem{NIPS2018_7964}
M.~Khayatkhoei, M.~K. Singh, and A.~Elgammal, ``Disconnected manifold learning
  for generative adversarial networks,'' in \emph{Advances in Neural
  Information Processing Systems 31}, S.~Bengio, H.~Wallach, H.~Larochelle,
  K.~Grauman, N.~Cesa-Bianchi, and R.~Garnett, Eds.\hskip 1em plus 0.5em minus
  0.4em\relax Curran Associates, Inc., 2018, pp. 7343--7353.

\bibitem{8627945}
C.~{Wang}, C.~{Xu}, X.~{Yao}, and D.~{Tao}, ``Evolutionary generative
  adversarial networks,'' \emph{IEEE Transactions on Evolutionary Computation},
  pp. 1--1, 2019.

\bibitem{NIPS2014_5423}
I.~Goodfellow, J.~Pouget-Abadie, M.~Mirza, B.~Xu, D.~Warde-Farley, S.~Ozair,
  A.~Courville, and Y.~Bengio, ``Generative adversarial nets,'' in
  \emph{Advances in Neural Information Processing Systems 27}, Z.~Ghahramani,
  M.~Welling, C.~Cortes, N.~D. Lawrence, and K.~Q. Weinberger, Eds.\hskip 1em
  plus 0.5em minus 0.4em\relax Curran Associates, Inc., 2014, pp. 2672--2680.

\bibitem{8681243}
Y.~{Tian}, X.~{Zheng}, X.~{Zhang}, and Y.~{Jin}, ``Efficient large-scale
  multiobjective optimization based on a competitive swarm optimizer,''
  \emph{IEEE Transactions on Cybernetics}, pp. 1--13, 2019.

\bibitem{Cheng2017Test}
R.~Cheng, Y.~Jin, M.~Olhofer, and B.~Sendhoff, ``Test problems for large-scale
  multiobjective and many-objective optimization,'' \emph{IEEE Transactions on
  Cybernetics}, vol.~47, no.~12, pp. 4108--4121, 2017.

\bibitem{7544478}
X.~{Zhang}, Y.~{Tian}, R.~{Cheng}, and Y.~{Jin}, ``A decision variable
  clustering-based evolutionary algorithm for large-scale many-objective
  optimization,'' \emph{IEEE Transactions on Evolutionary Computation},
  vol.~22, no.~1, pp. 97--112, Feb 2018.

\bibitem{Yang2014Large}
Z.~Yang, K.~Tang, and X.~Yao, ``Large scale evolutionary optimization using
  cooperative coevolution,'' \emph{Information Sciences}, vol. 178, no.~15, pp.
  2985--2999, 2014.

\bibitem{s7091892}
R.~{Shang}, K.~{Dai}, L.~{Jiao}, and R.~{Stolkin}, ``Improved memetic algorithm
  based on route distance grouping for multiobjective large scale capacitated
  arc routing problems,'' \emph{IEEE Transactions on Cybernetics}, vol.~46,
  no.~4, pp. 1000--1013, April 2016.

\bibitem{Bergh2004A}
F.~V.~D. Bergh and A.~P. Engelbrecht, ``A cooperative approach to particle
  swarm optimization,'' \emph{IEEE Trans.evol.comput}, vol.~8, no.~3, pp.
  225--239, 2004.

\bibitem{zille2017framework}
H.~Zille, H.~Ishibuchi, S.~Mostaghim, and Y.~Nojima, ``A framework for
  large-scale multiobjective optimization based on problem transformation,''
  \emph{IEEE Transactions on Evolutionary Computation}, vol.~22, no.~2, pp.
  260--275, 2017.

\bibitem{He2020Evolutionary}
H.~Cheng, S.~Huang, R.~Cheng, K.~C. Tan, and Y.~Jin, ``Evolutionary
  multiobjective optimization driven by generative adversarial networks
  (gans),'' \emph{IEEE Transactions on Cybernetics}, 2020.

\bibitem{dong2018musegan}
H.-W. Dong, W.-Y. Hsiao, L.-C. Yang, and Y.-H. Yang, ``Musegan: Multi-track
  sequential generative adversarial networks for symbolic music generation and
  accompaniment,'' in \emph{Thirty-Second AAAI Conference on Artificial
  Intelligence}, 2018.

\bibitem{ccfinproceedings}
F.~Chao, J.~Lv, D.~Zhou, L.~Yang, C.-M. Lin, C.~Shang, and C.~Zhou,
  ``Generative adversarial nets in robotic chinese calligraphy,'' in \emph{2018
  IEEE International Conference on Robotics and Automation (ICRA)}.\hskip 1em
  plus 0.5em minus 0.4em\relax IEEE, 2018, pp. 1104--1110.

\bibitem{NIPS2017_7159}
I.~Gulrajani, F.~Ahmed, M.~Arjovsky, V.~Dumoulin, and A.~C. Courville,
  ``Improved training of wasserstein gans,'' in \emph{Advances in Neural
  Information Processing Systems 30}.\hskip 1em plus 0.5em minus 0.4em\relax
  Curran Associates, Inc., 2017, pp. 5767--5777.

\bibitem{do2002wavelet}
M.~N. Do and M.~Vetterli, ``Wavelet-based texture retrieval using generalized
  gaussian density and kullback-leibler distance,'' \emph{IEEE transactions on
  image processing}, vol.~11, no.~2, pp. 146--158, 2002.

\bibitem{lin1991divergence}
J.~Lin, ``Divergence measures based on the shannon entropy,'' \emph{IEEE
  Transactions on Information theory}, vol.~37, no.~1, pp. 145--151, 1991.

\bibitem{zhang2004solving}
T.~Zhang, ``Solving large scale linear prediction problems using stochastic
  gradient descent algorithms,'' in \emph{Proceedings of the twenty-first
  international conference on Machine learning}.\hskip 1em plus 0.5em minus
  0.4em\relax ACM, 2004, p. 116.

\bibitem{10.2307/2346830}
J.~A. Hartigan and M.~A. Wong, ``Algorithm as 136: A k-means clustering
  algorithm,'' \emph{Journal of the Royal Statistical Society. Series C
  (Applied Statistics)}, vol.~28, no.~1, pp. 100--108, 1979.

\bibitem{UnsupervisedRepresentation}
A.~Radford and S.~Chintala, ``Unsupervised representation learning with deep
  convolutional generative adversarial networks,'' \emph{Computer Science},
  vol.~13, no.~2, pp. 284--302, 2015.

\bibitem{996017}
K.~{Deb}, A.~{Pratap}, S.~{Agarwal}, and T.~{Meyarivan}, ``A fast and elitist
  multiobjective genetic algorithm: Nsga-ii,'' \emph{IEEE Transactions on
  Evolutionary Computation}, vol.~6, no.~2, pp. 182--197, April 2002.

\bibitem{sainath2015convolutional}
T.~N. Sainath, O.~Vinyals, A.~Senior, and H.~Sak, ``Convolutional, long
  short-term memory, fully connected deep neural networks,'' in \emph{2015 IEEE
  International Conference on Acoustics, Speech and Signal Processing
  (ICASSP)}.\hskip 1em plus 0.5em minus 0.4em\relax IEEE, 2015, pp. 4580--4584.

\bibitem{7299173}
K.~{He} and J.~{Sun}, ``Convolutional neural networks at constrained time
  cost,'' in \emph{2015 IEEE Conference on Computer Vision and Pattern
  Recognition (CVPR)}, June 2015, pp. 5353--5360.

\bibitem{Tian2019Efficient}
Y.~Tian, X.~Zheng, X.~Zhang, and Y.~Jin, ``Efficient large-scale multiobjective
  optimization based on a competitive swarm optimizer,'' \emph{IEEE
  Transactions on Cybernetics}, pp. 1--13, 2019.

\bibitem{8065138}
Y.~{Tian}, R.~{Cheng}, X.~{Zhang}, and Y.~{Jin}, ``Platemo: A matlab platform
  for evolutionary multi-objective optimization [educational forum],''
  \emph{IEEE Computational Intelligence Magazine}, vol.~12, no.~4, pp. 73--87,
  Nov 2017.

\bibitem{10.1007/3-540-45356-3_83}
K.~Deb, S.~Agrawal, A.~Pratap, and T.~Meyarivan, ``A fast elitist non-dominated
  sorting genetic algorithm for multi-objective optimization: Nsga-ii,'' in
  \emph{Parallel Problem Solving from Nature PPSN VI}, M.~Schoenauer, K.~Deb,
  G.~Rudolph, X.~Yao, E.~Lutton, J.~J. Merelo, and H.-P. Schwefel, Eds.\hskip
  1em plus 0.5em minus 0.4em\relax Berlin, Heidelberg: Springer Berlin
  Heidelberg, 2000, pp. 849--858.

\bibitem{4633340}
H.~{Li} and Q.~{Zhang}, ``Multiobjective optimization problems with complicated
  pareto sets, moea/d and nsga-ii,'' \emph{IEEE Transactions on Evolutionary
  Computation}, vol.~13, no.~2, pp. 284--302, 2009.

\bibitem{ZHANG201863}
X.~Zhang, X.~Zheng, R.~Cheng, J.~Qiu, and Y.~Jin, ``A competitive mechanism
  based multi-objective particle swarm optimizer with fast convergence,''
  \emph{Information Sciences}, vol. 427, pp. 63 -- 76, 2018.

\bibitem{1197687}
E.~{Zitzler}, L.~{Thiele}, M.~{Laumanns}, C.~M. {Fonseca}, and V.~G. {da
  Fonseca}, ``Performance assessment of multiobjective optimizers: an analysis
  and review,'' \emph{IEEE Transactions on Evolutionary Computation}, vol.~7,
  no.~2, pp. 117--132, April 2003.

\bibitem{8605765}
G.~{Ismayilov} and H.~R. {Topcuoglu}, ``Dynamic multi-objective workflow
  scheduling for cloud computing based on evolutionary algorithms,'' in
  \emph{2018 IEEE/ACM International Conference on Utility and Cloud Computing
  Companion (UCC Companion)}, Dec 2018, pp. 103--108.

\bibitem{DERRAC20113}
J.~Derrac, S.~García, D.~Molina, and F.~Herrera, ``A practical tutorial on the
  use of nonparametric statistical tests as a methodology for comparing
  evolutionary and swarm intelligence algorithms,'' \emph{Swarm and
  Evolutionary Computation}, vol.~1, no.~1, pp. 3 -- 18, 2011.

\bibitem{8114198}
A.~{Gupta}, Y.~{Ong}, and L.~{Feng}, ``Insights on transfer optimization:
  Because experience is the best teacher,'' \emph{IEEE Transactions on Emerging
  Topics in Computational Intelligence}, vol.~2, no.~1, pp. 51--64, Feb 2018.

\bibitem{9122031}
M.~{JIANG}, Z.~{WANG}, H.~{HONG}, and G.~G. {YEN}, ``Knee point based
  imbalanced transfer learning for dynamic multi-objective optimization,''
  \emph{IEEE Transactions on Evolutionary Computation}, pp. 1--1, 2020.

\bibitem{8100935}
M.~{Jiang}, Z.~{Huang}, L.~{Qiu}, W.~{Huang}, and G.~G. {Yen}, ``Transfer
  learning-based dynamic multiobjective optimization algorithms,'' \emph{IEEE
  Transactions on Evolutionary Computation}, vol.~22, no.~4, pp. 501--514,
  2018.

\bibitem{9199822}
M.~{Jiang}, Z.~{Wang}, S.~{Guo}, X.~{Gao}, and K.~C. {Tan}, ``Individual-based
  transfer learning for dynamic multiobjective optimization,'' \emph{IEEE
  Transactions on Cybernetics}, pp. 1--14, 2020.

\bibitem{9002755}
G.~{CHI}, M.~{JIANG}, X.~{GAO}, W.~{HU}, S.~{GUO}, and K.~C. {TAN}, ``Online
  bagging for anytime transfer learning,'' in \emph{2019 IEEE Symposium Series
  on Computational Intelligence (SSCI)}, 2019, pp. 941--947.

\end{thebibliography}

\end{document}